\pdfoutput=1

\documentclass[11pt]{article}

\usepackage{scrextend}
\usepackage{emnlp2021}
\usepackage{times}
\usepackage{inconsolata}
\usepackage{latexsym}

\usepackage{microtype}

\usepackage[T1]{fontenc}
\usepackage[utf8]{inputenc}

\usepackage{comment}
\usepackage{pgfplots}
\usepackage{caption,subcaption}
\usepackage{adjustbox}

\usepackage{amsmath,amsfonts,amssymb}
\usepackage[noline,boxed,noend]{algorithm2e}
\usepackage{graphicx}

\usepackage{xcolor}

\usepackage{booktabs}
\usepackage{multicol}
\usepackage{multirow}

\usepackage{tikz}

\newcommand{\ra}[1]{\renewcommand{\arraystretch}{#1}}

\newcommand\daniel[1]{}
\newcommand\justin[1]{}

\newcommand\eg{e.g.,\ }
\newcommand\ie{i.e.,\ }

\newcommand{\Algorithm}[1]{Algorithm~\ref{#1}}

\newcommand{\Figure}[1]{Figure~\ref{#1}}
\newcommand{\Equation}[1]{Equation~\ref{#1}}

\newcommand{\Section}[1]{Section~\ref{#1}}
\newcommand{\Appendix}[1]{Appendix~\ref{#1}}

\DeclareMathOperator*{\argmax}{arg\,max}

\DeclareMathOperator*{\topk}{topk}

\newcommand{\zutt}{\ensuremath{u}}
\newcommand{\zdot}{\ensuremath{\mathbf{r}}}
\newcommand{\zref}{\ensuremath{\mathbf{r}}}
\newcommand{\zdotconfig}{\ensuremath{r}}
\newcommand{\zrefconfig}{\ensuremath{r}}
\newcommand{\zmen}{\ensuremath{\mathbf{r}}}

\newcommand{\zcon}{\ensuremath{c}}
\newcommand{\zwor}{\ensuremath{w}}
\newcommand{\zmem}{\ensuremath{M}}
\newcommand{\zsel}{\ensuremath{s}}

\newcommand{\zlistener}{\ensuremath{P_R}}
\newcommand{\zspeaker}{\ensuremath{P_U}}
\newcommand{\zmention}{\ensuremath{P_M}}
\newcommand{\zchoice}{\ensuremath{P_S}}

\newcommand{\onecommon}{\textsc{OneCommon}\xspace}

\newcommand{\cmention}{\ensuremath{\text{RNN}_M}\xspace}
\newcommand{\chistory}{\ensuremath{\text{RNN}_C}\xspace}

\newcommand{\mlp}{\text{MLP}}

\newcommand{\modelfull}{\textsc{Full}\xspace}
\newcommand{\modelprag}{\textsc{Full+Prag}\xspace}
\newcommand{\nomem}{\textsc{--Mem}}
\newcommand{\nostruc}{\textsc{--Struc}\xspace}

\title{Reference-Centric Models for Grounded Collaborative Dialogue}

\author{Daniel Fried$^\dagger$ \qquad Justin T.\ Chiu$^\ddagger$ \qquad Dan Klein$^\dagger$ \\
$^\dagger$Computer Science Division, UC Berkeley\\ 
$^\ddagger$Department of Computer Science, Cornell Tech \\
{\tt \{dfried,klein\}@cs.berkeley.edu, jtc257@cornell.edu}
}

\date{}

\begin{document}
\maketitle
\begin{abstract}
  We present a grounded neural dialogue model that successfully collaborates with people in a partially-observable reference game. We focus on a setting where two agents each observe an overlapping part of a world context and need to identify and agree on some object they share. Therefore, the agents should pool their information and communicate pragmatically to solve the task.  Our dialogue agent accurately grounds referents from the partner's utterances using a structured reference resolver, conditions on these referents using a recurrent memory, and uses a pragmatic generation procedure to ensure the partner can resolve the references the agent produces. We evaluate on the OneCommon spatial grounding dialogue task \citep{udagawa2019natural}, involving a number of dots arranged on a board with continuously varying positions, sizes, and shades. Our agent substantially outperforms the previous state of the art for the task, obtaining a 20\% relative improvement in successful task completion in self-play evaluations and a 50\% relative improvement in success in human evaluations. 

\end{abstract}

\section{Introduction}
\label{sec:introduction}

\begin{figure}[t]
  \centering
\includegraphics[width=0.72\linewidth]{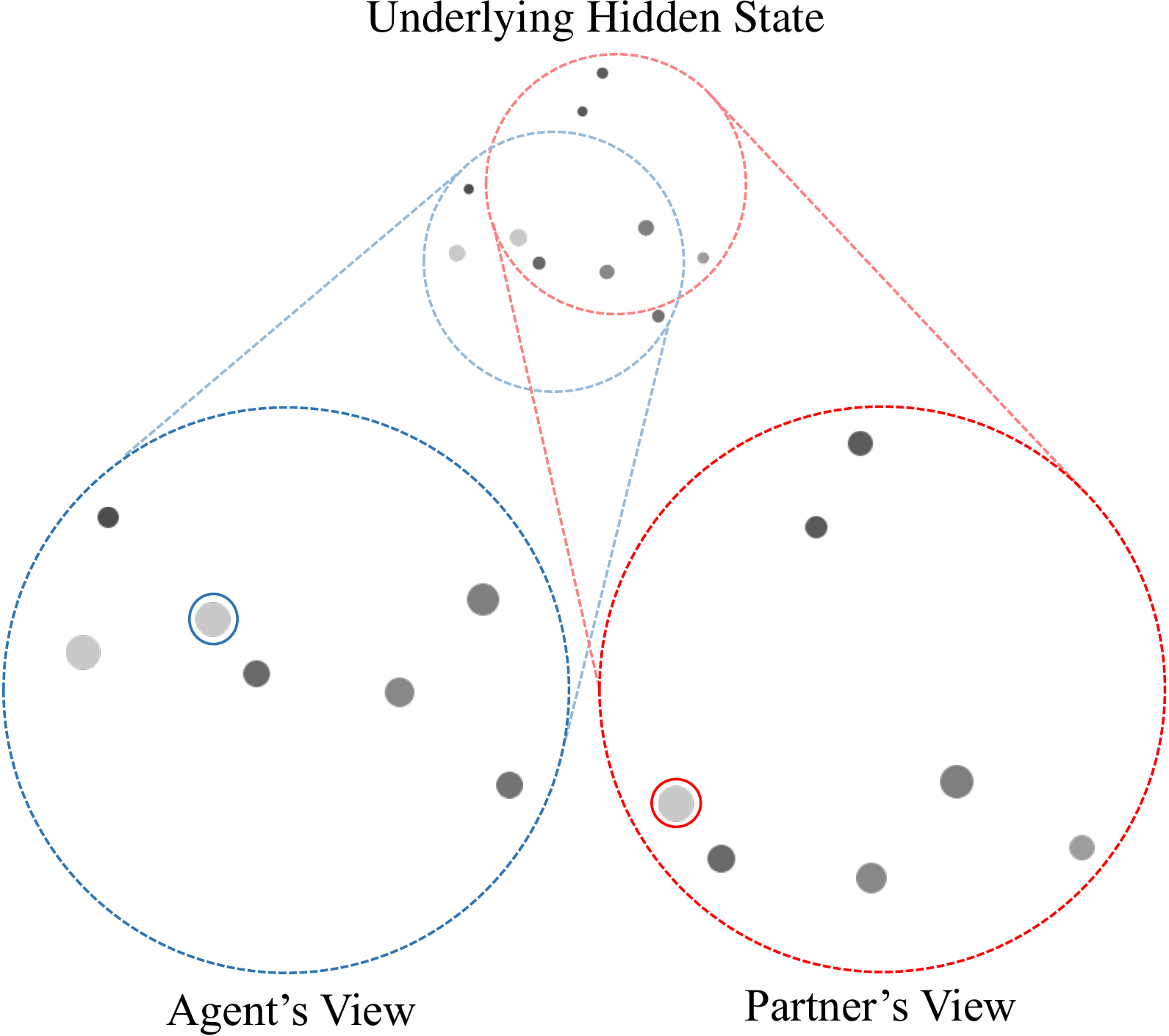}
\small
\begin{tabular}{@{}l@{}l@{}}
\toprule
{\color{blue} A}:\;&I have two large light grey dots with a smaller darker dot \\
&below and to the right of it \\
{\color{red} P}:\;&I have one large lightest grey dot in my entire view \\
{\color{blue} A}:\;&I have two large light grey dots \\
{\color{red} P}:\;&There is a smaller slightly darker grey dot next to the\\ 
&lightest grey and larger dot \\
{\color{blue} A}:\;&Yes , let's pick the light one \\
{\color{red} P}:\;&The light grey and large dot? \\
{\color{blue} A}:\;&Yes it 's the one . Let's pick that one \\
{\color{red} P}:\;&SELECT {\color{red} red} \\
{\color{blue} A}:\;&SELECT {\color{blue} blue} \\
\bottomrule
\end{tabular}
\caption{
An example dialogue produced by our system ({\color{blue} A}) with a human partner ({\color{red} P}).
The participants have different but overlapping views of a shared board,
which contains dots of different shapes and sizes.
The partners must collaborate through dialogue in order to find and select a dot that is visible to both.
\vspace{-1.8em}
}
\label{fig:dialogue-example} 
\end{figure}

In grounded dialogue settings involving high degrees of ambiguity, correctly interpreting and informatively generating language can prove challenging.
Consider the collaborative dialogue game shown in Figure~\ref{fig:dialogue-example}.
Each player has a separate, but overlapping, view on an underlying context. 
They need to communicate to determine and agree on one dot that they share, and both players win if they choose the same dot.
To succeed, each participant must---implicitly or explicitly---ground their partner's descriptions to their own context, maintain a history of what's been described and what their partner is likely to have, and informatively convey parts of their own context.

We present a grounded pragmatic dialogue system which collaborates successfully with people on the task above. Figure~\ref{fig:dialogue-example} shows a real example game between our system and a human partner.
Our approach is centered around a structured module for perceptually-grounded reference resolution. 
This reference resolution module plays two roles. First, the module is used to \emph{interpret} the partner's utterances: explicitly predicting which referents (if any) in the agent's context the partner is referring to, for example \emph{a smaller darker grey dot} and \emph{the lightest grey and larger dot}. 
Second, the reference module is used for \emph{pragmatic generation}: choosing utterances by reasoning about how the partner might interpret them in context. Our pragmatic generation procedure selects referents to describe as well as choosing how to describe them, for example focusing on \emph{the light one} (\Figure{fig:dialogue-example}).

Much past work that has constructed systems for grounded collaborative dialogue has focused on settings that have asymmetric player roles
\citep{kim-etal-2019-codraw,devries2018talk,Das_2018_CVPR,visdial}, are fully-observable, or are grounded in symbolic attributes \citep{he2017mutualfriends}.  
In contrast, we focus on the \onecommon corpus and task \cite{udagawa2019natural}, which is symmetric, partially-observable, and has relatively complex spatial and perceptual grounding. 
These traits necessitate complex dialogue strategies such as common grounding, coordination, clarification questions, and nuanced acknowledgment \citep{udagawa2019natural},
leading to the task being challenging even for pairs of human partners.

Past work on \onecommon has 
focused on the subtask of reference resolution \cite{udagawa2020annotated,udagawa2020spatial} and only evaluated dialogue systems automatically: using static evaluation on human--human games and self-play evaluations that simulate human partners using another copy of the agent.
Our system outperforms this past work on these evaluations.
We further confirm these results by performing---for the first time on this task---human evaluations, where we find that our system obtains a 50\% relative increase in success rate over a system from past work when paired with human partners. We release code for our system at \texttt{\href{https://github.com/dpfried/onecommon}{https://github.com/dpfried/onecommon}}.

\section{Setting}
\label{sec:setting}
We choose to focus on the \onecommon task \citep{udagawa2019natural} since it is a particularly challenging representative of a class of partially-observable collaborative reference dialogue games (\eg \citealt{he2017mutualfriends,haber2019photobook}).
In this task, two players have different but overlapping views of a game board, which consists of dots of various positions, shades of gray and sizes. The players must coordinate to choose a single dot that both players can see, which is challenging because neither knows which dots the other can see. 

Each player's world view, $w$, consists of a circular view on an underlying board containing between 8 and 10 randomly scattered dots, with continuously varying positions, shades, and sizes (Figure~\ref{fig:dialogue-example}). 
Each player's view contains 7 dots, and the views of the players
overlap so that there are between 4 and 6 dots which appear in both views.

We focus on a turn-based version of the dialogue task. 
In a given turn $t$, a player may communicate with their partner by either sending an utterance $u_{t}$ or selecting a dot $s$. In the event of selection, the partner is notified but cannot see which dot the player has selected.
Once a player has selected a dot, they can no longer send messages.
The dialogue ends once both players have selected a dot, and is successful if both selected the same one.

\section{Model Structure}
\label{sec:model}
\begin{figure*}
    \centering
    \includegraphics[width=0.99\textwidth]{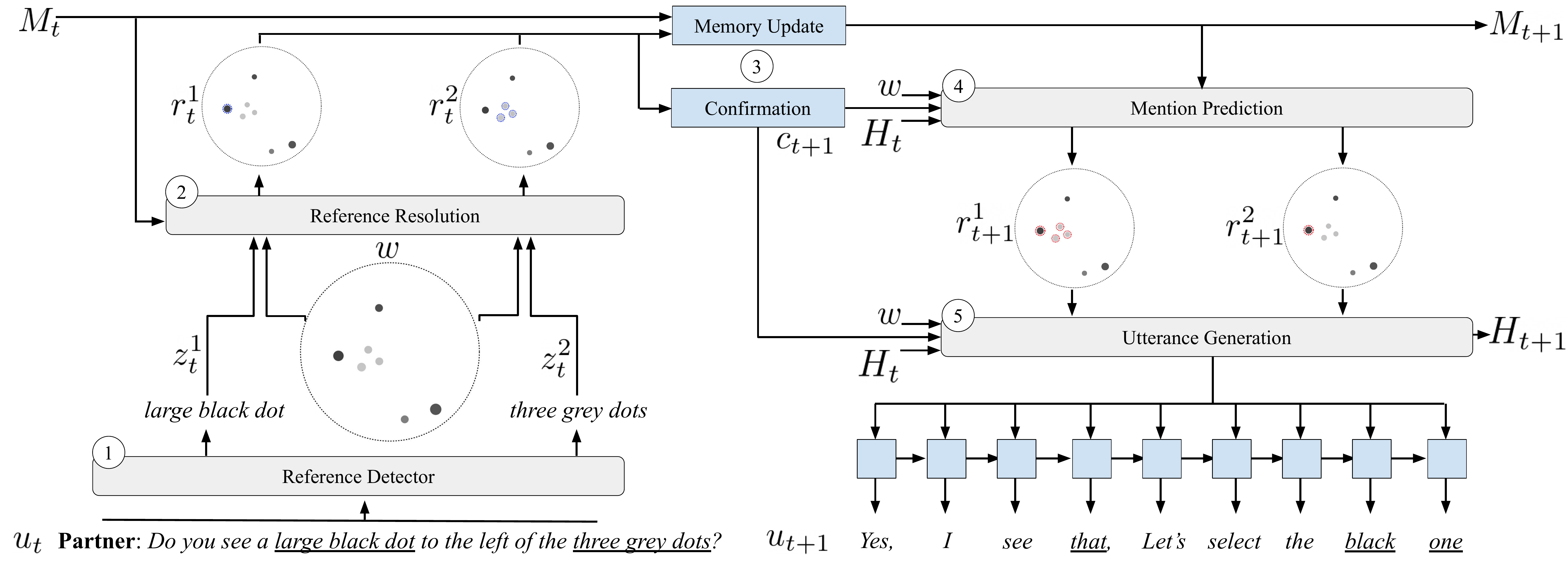}
    \caption{
    \label{fig:turn-process}
    In a given turn, an agent first identifies referring expressions in their partner's utterance $u_t$ using the reference detector (1). Each reference is then resolved with the reference resolution module (2), which uses encoded representations $z^{1:K_{t}}$ of the reference segments and the world context $\zwor$.  The referents are then used to update the referent memory $M_t$, and cross-referenced against the agent's own dots to confirm whether or not the agent can also see them (3). Given the referent memory $M_t$ and confirmation variable $c_{t+1}$, the mention prediction module (4) produces a sequence of dot configurations $z_{t+1}^{1:K_{t+1}}$ to mention. Finally, the utterance generation module (5) uses the dialog history $H_t$, confirmation variable, and attended representations of the selected mentions and world context to generate a response $u_{t+1}$.
    \vspace*{-1.4em}
  }
\end{figure*}

Our approach is a modular neural dialogue model which factors the agent's generation process into a series of successive subtasks, all centered on grounding language into referents in the world context. %
In this section, we describe our model structure, which defines a neural module for each subtask.
We then describe our reference-centric pragmatic generation procedure in \Section{sec:pragmatics}.

An overview of the relationship between modules in our model is shown in Figure~\ref{fig:turn-process}.
Each module can condition on neural encodings of the context (the world and past dialogue), as well as the outputs of other modules. 
We describe our system at a high-level here, then give task-specific implementation details about each component in \Section{sec:implementation}.

\subsection{Context Encodings} 
Our modules can condition on encodings of (i) the past utterances $u_{1:t}$ in the dialogue, represented as a memory vector $H_t$ produced by a word-level recurrent encoder
and (ii) the continuous dots in the world context $\zwor$, produced by the entity encoding network of \citet{udagawa2020annotated}, which produces a vector $w(d)$ for each dot $d$ encoding the dot's continuous attributes as well as its pairwise attribute relationships to all other dots in the context \cite{santoro2017}.
(i) and (ii) both follow \citet{udagawa2020annotated}. 
To explicitly encourage the model to retain and use information about the history of referents mentioned by both players, which affects the choice of future referents as well as the selection of dot at the end of the game, we also use (iii) a structured recurrent \textbf{referent memory} grounded in the context. 
This memory, inspired by \citet{he2017mutualfriends}, has one representation for each dot $d$ in the agent's view, $M_t(d)$, which is updated based on the referents predicted in turn $t$. See \Section{sec:referent-memory} for details.

\subsection{Decomposing Turns into Subtasks}
\label{sec:decomposing-turns}
We assume turn $t+1$ in the dialogue has the following generative process (numbers correspond to Figure~\ref{fig:turn-process}).
Steps (1) and (2) identify and resolve referring expressions in the partner's utterance $u_t$; step (3) updates the memory and determines whether the model can confirm any referents from the partner's utterance; steps (4) and (5) produce the agent's next utterance $u_{t+1}$.
\\(1) First, a sequence of $K_t$ (with $K_t\ge0$) referring expressions are identified in $u_t$ using the \textbf{reference detector} tagging model of \citet{udagawa2020annotated}\footnote{Udagawa and Aizawa refer to this as a \emph{markable detector} given their work's focus on referent annotation.}, and encodings $\mathbf{z}_t = z_{t}^{1:K_t}$ are obtained for them by pooling features from a recurrent utterance encoder. 
\\ (2)
  Then, the referring expressions are grounded. From each referring expression's features $z^{k}$, we predict a referent $r^k$, which is the set of zero or more dots in the agent's own view which are described by the referring expression.
For example, the referring expression \emph{\underline{three gray dots}} corresponds to a single referent containing three dots. 
A \textbf{reference resolution} module $\zlistener(\zref_{t} \mid \mathbf{z}_t, \zwor, M)$, where $\zref_{t} = \zrefconfig_{t}^{1:K_t}$, predicts a sequence of referents, one for each referring expression. %
\\ (3) Given these referents, the agent updates the referent memory $M_t$ using the predicted referents and constructs a discrete \textbf{confirmation variable} $\zcon_{t+1}$, which indicates whether the agent can confirm in its next utterance that it has all the referents the partner is describing (\eg \emph{Yes, I see that}).  $\zcon_{t+1}$ takes on one of three values: \textsc{NA} if no referring expressions were in the partner's utterance, \textsc{Yes} if all of the partner's referring expressions have referents that are at least partially visible in the agent's view, and \textsc{No} otherwise.
\\(4) The agent chooses a sequence of referents to mention next using a \textbf{mention prediction} module $\zmention(\zmen_{t+1} \mid c_{t+1}, \zmem_{t+1}, H_t, \zwor)$. %
 \\ (5) Finally, the next utterance $\zutt_{t+1}$ is produced using an \textbf{utterance generation} module $\zspeaker(\zutt_{t+1}\mid \zmen_{t+1},c_{t+1}, H_t, w)$, also updating the word-level recurrent memory $H_{t+1}$. %

At the end of the dialogue (turn $T$), the agent selects a dot $s$ using a \textbf{choice selection} module $\zchoice(s\mid H_T,M_T,\zwor)$ (not shown in Figure~\ref{fig:turn-process}).\footnote{The choice selection module is invoked when the utterance generation model predicts a special \textsc{<SELECT>} token, following \citet{udagawa2020annotated}.}

Modules that predict referents (reference resolution, mention selection, and choice selection) are all implemented using a structured conditional random field (CRF) architecture (\Section{sec:ref_resolution}), with independent parameterizations for each module.

Our model bears some similarities to \citet{udagawa2020annotated}'s neural dialogue model for this task: both models use a reference resolution module\footnote{Our model, however, uses a structured CRF while Udagawa and Aizawa's model does not use structured output modeling.} and both models attend to similar encodings of the dots in the agent's world view ($w(d)$) when generating language.
Crucially, however, our decomposition of generation into subtasks results in a factored, hierarchical generation procedure: our model identifies and then conditions on previously-mentioned referents from the partner's utterances,\footnote{Udagawa and Aizawa used the reference resolution module only to define an auxiliary loss at training time.} maintains a structured referent memory updated at each utterance, and explicitly predicts which referents to mention in each of the agent's own utterances. In \Section{sec:pragmatics}, we will see how factoring the generation procedure in this way allows us to use a pragmatic generation procedure, and in \Section{sec:experiments} we find that each of these components improves performance.

\section{Pragmatic Generation}
\label{sec:pragmatics}

The modules as described above can be used to generate the next utterance $u_{t+1}$ using the predictions of $\zmention(\zmen_{t+1})$ and $\zspeaker(\zutt_{t+1} | \zmen_{t+1})$ (omitting other conditioning variables from the notation for brevity; see \Section{sec:model} for the full conditioning contexts).
This section describes an improvement, \emph{pragmatic generation}, to this process.
Referents and their expressions should be relevant in the dialogue and world context, but they should also be discriminative \cite{dale1989cooking}: allowing the listener to easily understand which referents the speaker is intending to describe.
Our pragmatic generation approach, based on the Rational Speech Acts (RSA) framework \cite{Frank12predictingpragmatic,GoodmanFrank2016-TICS}, uses the reference resolution module, $\zlistener(\zmen_{t+1} | \zutt_{t+1})$, to predict whether the partner can identify the intended referents. This encourages selecting referents that are easy for the partner to identify and describing them informatively in context.\footnote{Note that the reference resolution model, which has access to the agent's own view and not the partner's, can only approximate whether the referents are identifiable by the partner; nevertheless we find that it is beneficial for pragmatic generation. Future work might explore also inferring and using the partner's view.}

We use the following objective over referents $\zdot$ and utterances $\zutt$ for a given turn:
\begin{align}
  \label{eqn:pragmatic-objective}
\argmax_{\zdot, \zutt} ~ & L(\zdot, \zutt) \nonumber \\
L(\zdot, \zutt) &= \zmention(\zdot)^{w_M} \cdot \zspeaker(\zutt|\zdot)^{w_S} \cdot \zlistener(\zdot|\zutt)^{w_L} 
\end{align}
where $w_M$, $w_S$, and $w_L$ are hyperparameters. %

This objective generalizes the typical RSA setup (as implemented by the weighted pragmatic inference objective of \eg \citealt{andreas2016reasoning} and \citealt{monroe-etal-2017-colors}), which chooses \emph{how} to describe a given context (\ie choosing an utterance $u$), to also choose \emph{what} context to describe (\ie choosing the referents $\zdot$).
Our objective also models the tradeoff, explored in past work on referring expression generation \cite{dale1989cooking,jordan2005contentselection,viethen2011generating}, between producing utterances relevant in the discourse and world context and producing utterances that are discriminative. We use $P_U$ and $P_M$ to model discourse and world relevance, $P_S$ to model discriminability, and the weights $w$ to empirically model the tradeoff between them.

\begin{figure}[t]
    \centering
    \includegraphics[width=1.06\linewidth]{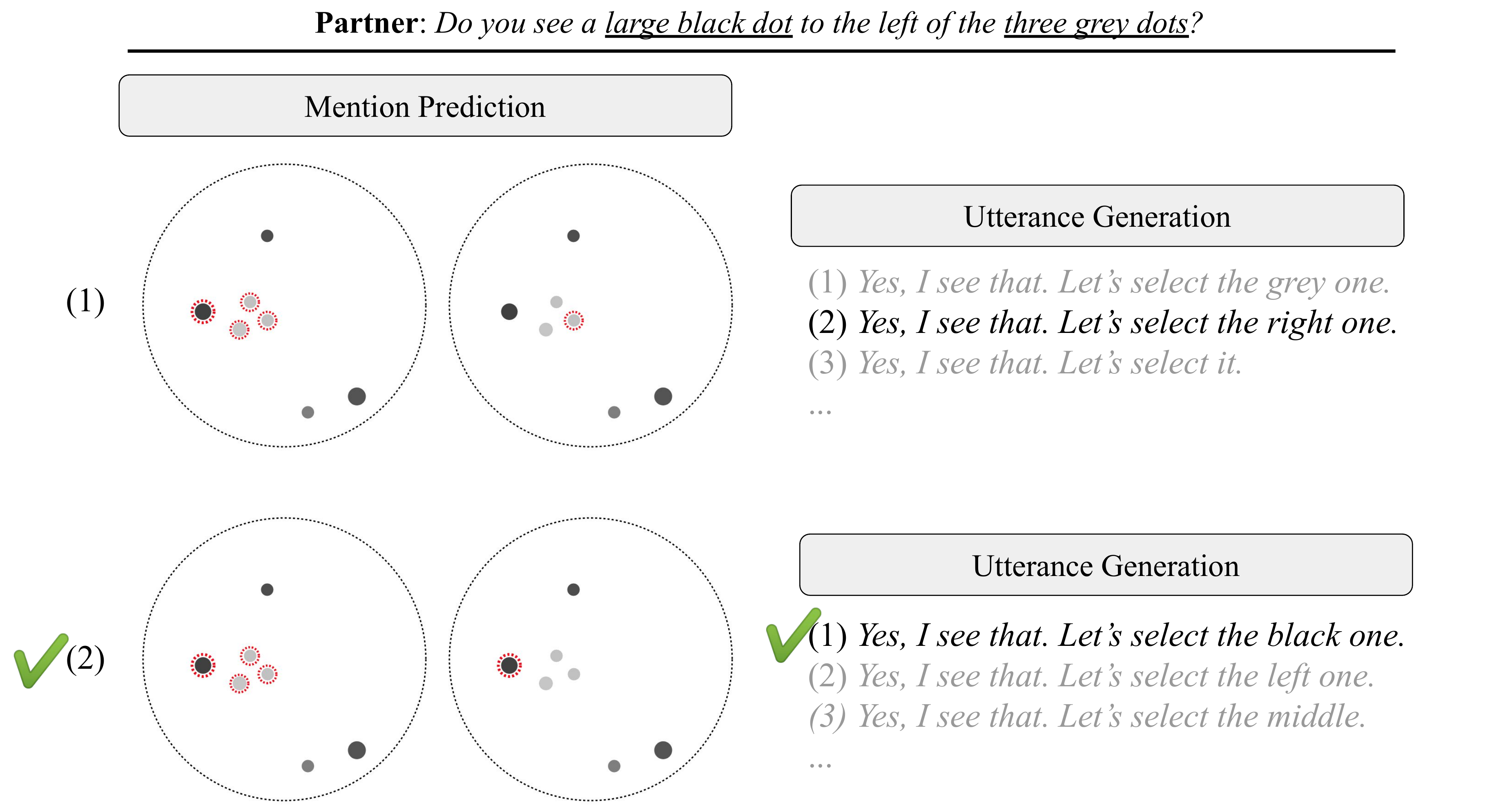}
    \caption{
    \label{fig:prag-gen}
    Agents optimize for a combination of fluency and informativity during pragmatic utterance generation (\Section{sec:pragmatics} and \Algorithm{alg:praggeneration}).
      A set of paired candidate referents (from the mention prediction module) and utterances (from the utterance generation module) is rescored using $L(\zdot,\zutt)$ (\Equation{eqn:pragmatic-objective}), a weighted geometric mean of scores from the mention prediction, utterance, and reference resolution modules. %
The pair of referent and utterance that maximizes this score is chosen as a response.
\vspace*{-1em}
}
\end{figure}

Given the combinatorially-large spaces of possible $\zdot$ and $\zutt$, we rely on an early-stopping approximate search, which to our knowledge is novel for RSA. The search (illustrated in \Figure{fig:prag-gen}) iterates through the highest probability structured referent sequences $\zdot$ under the mention prediction module $\zmention$ (\Figure{fig:prag-gen} shows the top two) and for each $\zdot$ sampling $N_\zutt$ utterances $\zutt$ from the utterance generation module (\Figure{fig:prag-gen} shows three $\zutt$ per $\zdot$). If the maximum of these $(\zdot, \zutt)$ pairs under $L$ is better than an early-stopping threshold value $\tau$, we return the pair. Otherwise, we continue on to the next $\zdot$. If more than $N_r$ referent sequences have been evaluated, we return the best $(\zdot, \zutt)$ pair found so far. See \Appendix{app:pragmatic-generation} for pseudocode and a discussion of robustness to the threshold $\tau$. %

\section{Module Implementations}
\label{sec:implementation}
As described so far, our system is applicable to a range of partially-observable grounded collaborative referring expression dialogue tasks (\eg \citealt{he2017mutualfriends,haber2019photobook}). In this section, we describe implementations of our systems' modules, some of which are tailored to \onecommon.

\subsection{Reference Detection}
We identify a sequence of referring expressions in the utterance using the \textbf{reference detector}
of \citet{udagawa2020annotated}, a BiLSTM-CRF tagger \citep{huang-2015-lstm-crf}.
Then, following \citet{udagawa2020annotated}, we obtain features $z^k$ for each of the $K$ referring expressions in the utterance (for use in the reference resolution model) with a bidirectional recurrent encoder, using learned weights to pool the encodings at the referring expression's boundaries as well as the end of the utterance.\footnote{The bidirectional encoder only has access to the utterances that have been produced so far, \ie $u_{1:t}$ when the agent is generating utterance $u_{t+1}$.}

\subsection{Structured Reference Resolution}
\label{sec:ref_resolution}
We use a structured reference resolution module to ground the referring expressions identified above: %
identifying dots in the agent's own view described by each expression. %
Grounding referents in this domain involves reasoning not only about attributes of individual dots 
but also spatially and comparatively within a single referring expression (\eg \emph{\underline{a line of three dots}}) or across referring expressions (\eg \emph{\underline{a large grey dot} left of \underline{a smaller dot}}).

To predict a sequence of referents $\zdot = r^{1:K}$ from the $K$ referring expression representations $z^{1:K}$ extracted above, we use a linear-chain CRF \cite{lafferty2001conditional} with neural potentials to parameterize
$\zlistener(\zdotconfig^{1:K} | z^{1:K}, \zwor, \zmem)$.
This architecture generalizes the reference resolution and choice selection models of \citet{udagawa2020annotated} and \citet{udagawa2020spatial} to model, in the output structure, relationships between dots, both inside and across referring expressions. %

There are three different types of potentials, designed to model language-conditioned features of individual dots $d$ in a referent $r$, $\phi$; relationships within a referent, $\psi$, and transitions between successive referents, $\omega$.
Given these potentials, the distribution is parameterized as
\par\nobreak
\vspace{-1em}
{\small
  \setlength{\abovedisplayskip}{6pt}
  \setlength{\belowdisplayskip}{\abovedisplayskip}
  \setlength{\abovedisplayshortskip}{0pt}
  \setlength{\belowdisplayshortskip}{5pt}
\begin{align*}
&P(\zdotconfig^{1:K} | z^{1:K}) \propto \\
&\exp\left(\sum_{k} f(\zdotconfig^k, z^k) + \psi(\zdotconfig^k,z^k) + \omega(\zdotconfig^{k:k+1},z^{k:k+1})\right),
\end{align*}
}
where $f(\zdotconfig, z) = \sum_{d \in r} \phi(d, z)$, and we've omitted the dependence of all terms on $\zmem$ and $\zwor$ for brevity.
We share all module parameters across the two subtasks of resolving referents for the agent and for the partner.%
\footnote{Parameters are shared for efficiency; sharing had little effect on performance in preliminary experiments.}

\paragraph{Individual Dots.} 
Dot potentials $\phi$ model the correspondence between language features $z^k$
and individual dots represented by encodings $\zwor(d)$, as well as discourse salience using the dot-level memory $\zmem(d)$ that tracks when the dot $d$ has been mentioned:\footnote{For the subtasks of reference resolution and choice selection, dot potentials are the same as the \emph{attention module} used by \citet{udagawa2020annotated}, with the addition of the memory $\zmem$.}
$$\phi(d, z^k) = \mlp_\phi([\zmem(d), z^k, \zwor(d)])$$

\paragraph{Dot Configurations.}

Configuration potentials $\psi(r^k, z^k)$ model the correspondence between language features and the set of all active dots in the agent's view for a referent $r^k$. 
These potentials further decompose into (1) pairwise potentials between active dots in the configuration, which relate the language embedding $z^k$ to attribute differences between dots in the pair (including as relative position, size, and shade) and (2) a potential on the entire configuration, which relates the language embedding to an embedding for the count of active dots in the configuration. 
See \Appendix{app:structured-crf} for more detail.

\paragraph{Configuration Transitions.}
Transition potentials $\omega(r^k, r^{k+1}, z^k, z^{k+1})$ model the correspondence between language features and relationships between referring expressions, \eg \emph{to the left of} in \emph{\underline{the black dot} to the left of \underline{the triangle of gray dots}}.
See \Appendix{app:structured-crf} for details.

\subsection{Confirmations}
When applied to the partner's utterances, the reference resolution module gives a distribution over which referents the \emph{partner} is likely to be referring to in the \emph{agent's} own context. 
If the agent can identify the referents its partner is describing, it should be able to confirm them, both in the dots it talks about next (\eg choosing to refer to one of the same dots the partner identified) and in the text of the utterances (\eg \emph{yes, I see it}). 
The discrete-valued confirmation variable (defined in \Section{sec:model}) models this, taking the value
\textsc{NA} if no referring expressions were identified in the partner's utterance, \textsc{Yes} if all of the $K>0$ referring expressions have a non-empty referent (at least one dot predicted in the agent's context) and \textsc{No} otherwise.

\subsection{Referent Memory}
\label{sec:referent-memory}
The memory state is composed of one state vector $M_t(d)$ for each dot in the agent's own context.
These dot states are updated using the referents identified in each utterance. This update is parameterized using a decoder cell, which is applied separately to each dot state:
\begin{align*}
\zmem_{t+1}(d) &= \chistory(\zmem_{t}(d), \iota(d, \zdot_{t}))
\end{align*}
where $\iota$ is a function that extracts features from the predictive distribution over referents from the previous utterance, representing mentions of dot $d$ in the referring expressions. We implement the cell using a GRU \cite{cho-etal-2014-learning}. See \Appendix{app:referent-memory} for more details.

\subsection{Mention Selection}
\label{sec:mention_prediction}
The mention selection subtask requires predicting a sequence of referents to mention in the agent's next utterance, $\zmention(\zdot_{t+1} \mid \zutt_{1:t}, \zmem_{t+1}, \zcon_{t+1}, \zwor)$.
To produce these referents, we use the same structured CRF architecture as the reference resolution module $\zlistener$. However, we use separate parameters from that module, and instead of the 
referring-expression inputs $\mathbf{z}$ use a sequence of vectors $x^{1:K_{t+1}}$ produced by
a decoder cell $\text{RNN}_M$, implemented using a GRU \cite{cho-etal-2014-learning}. %
The decoder conditions on
the dialogue context representation $H_t$ from the end of the last utterance, %
a learned vector embedding for the confirmation variable $\zcon_{t+1}$, and a mean-pooled representation of the memory $m = \frac{1}{|d|} \sum_d \zmem(d)$:
\begin{align*}
x^k &= \cmention(x^{k-1}, [H_t, \zcon_{t+1}, m])
\end{align*}
We obtain the number of referents $K_{t+1}$ by predicting at each step $k$ whether to halt from each $x^k$ using a linear layer followed by a logistic function. 

\subsection{Choice Selection}
\label{sec:choice_selection}
To parameterize the choice selection module $\zchoice(\zsel \mid \zutt_{1:T}, \zmem_T, \zwor)$, 
we again reuse the CRF architecture, with independent parameters from reference resolution and mention selection modules, replacing reference resolution's inputs $z^{1:K}$ with the dialogue context representation $H_T$ from the end of the final utterance in the dialogue. Since only a single dot needs to be identified, we use only the CRF's individual dot potentials $\phi$, removing $\psi$ and $\omega$. 
This is equivalent to the choice selection model (\textsc{TSEL}) used by \citet{udagawa2020annotated} if the recurrent memory $\zmem_T$ is removed.

\subsection{Utterance Generation}
\label{sec:utterance_module}
The utterance generation module $\zspeaker(\zutt_{t+1} | \zdot_{t+1}, c_{t+1}, H_t, w)$ is a sequence-to-sequence model. The module first encodes the sequence 
of dot encodings $w(d)$ for dots in the referents $z^{1:K_{t+1}}_{t+1}$ (predicted by the mention selection module) to produce encodings $y_{t}^{1:K}$.
Words in the utterance are then produced one at a time using a recurrent decoder
that has a hidden state initialized with 
a function that combines $y_{t}^{1:K}$, the dialog context $H_t$, and a learned embedding for the discrete confirmation variable $c_{t+1}$. 
The decoder has two attention
mechanisms over: (i) dot encodings $\zwor(d)$, following \citet{udagawa2020annotated}, and (ii) the sequence of encoded referents $y_{t}^{1:K_{t+1}}$. 
See \Appendix{app:utterance-generation} for details.

\section{Experiments}
\label{sec:experiments}

We compare our approach to past systems for the \onecommon dataset.
While our primary evaluation is to evaluate systems on their success rate on the full dialogue game when paired with human partners (\Section{sec:human_evaluation}), we also compare our system to past work, and ablated versions of our full system, using the automatic evaluations of past work.
\subsection{Models}
We compare our full system (\modelfull) to ablated versions of it that successively remove: (i) the referent memory, ablating explicit tracking of referents mentioned (F\nomem) and (ii) the structured potentials $\psi, \gamma$ in the reference resolution and mention selection modules (F\nomem\nostruc), removing explicit modeling of relationships within and across referents.
We also compare to a reimplementation of the system of \citet{udagawa2020annotated}, which we found obtained better performance than their reported results in all evaluation conditions due to implementation improvements.  See \Appendix{app:hyperparameters}.

We obtain supervision for all components of the systems by training on the referent-annotated corpus of 5,191 successful human--human dialogues collected by Udagawa and Aizawa (\citeyear{udagawa2019natural,udagawa2020annotated}).
See \Appendix{app:training-details} for training details.
We train one copy of each model on each of the corpus's 10 cross-validation splits. %
We report means and standard deviations across the splits' models, except in human evaluations where we use a single model.

\subsection{Corpus Evaluation}
\label{sec:corpus_evaluation}
Following \citet{udagawa2020annotated}, we evaluate models' accuracy at (1) predicting the dot chosen at the end of the game (Choice Acc.) using $\zchoice$ and (2) resolving the referents in utterances from the human partner in the dialogue who had the agent's view (Ref Resolution dot-level accuracy Acc.\ and exact match accuracy Ex.) using $\zlistener$.  
\begin{table}
\begin{adjustbox}{width=0.9\columnwidth,center}
\begin{tabular}{@{}lccc@{}} 
\toprule
& Choice & \multicolumn{2}{c}{Ref. Resolution} \\
Model  &  Acc. & Acc. & Ex.\\
\midrule
U\&A (2020) & 69.3$\pm$2.0 & 86.4$\pm$0.4 & 35.0$\pm$2.0 \\
U+ (2020)& -- & 86.0$\pm$0.3 & 54.9$\pm$0.8  \\
\hline
F\nomem\nostruc & 71.6$\pm$0.9 & 87.7$\pm$0.2 & 44.3$\pm$0.5 \\
F\nomem& 70.9$\pm$1.2 & 92.6$\pm$0.2 & 76.2$\pm$0.5 \\
\modelfull & 83.3$\pm$1.2 & 93.3$\pm$0.2 & 78.2$\pm$0.5 \\
\midrule
Human & 90.8 & 96.3 & 86.9 \\
\bottomrule
\end{tabular}
\end{adjustbox}
\caption{\label{tbl:static_results}
Accuracies for predicting the dot selected at the end of the game (Choice Acc.) and resolving referents from utterances produced in the agent's own perspective (dot-level accuracy Acc. and exact match Ex.) in 10-fold cross-validation on the corpus. Our \modelfull model outperforms all past work on the dataset: 
U\&A (2020) is our reimplementation of \citet{udagawa2020annotated}, and U+ (2020) are taken from \citet{udagawa2020spatial}. Human scores are annotator agreements \cite{udagawa2019natural}.
\vspace{-1.5em}
}
\end{table}

We see in Table \ref{tbl:static_results} that our \modelfull model improves substantially on past work, including the work of \citet{udagawa2020spatial}, who augment their referent resolution model with numeric features. Our structured reference resolver is able to learn these features in its potentials $\psi$ (in addition to other structured relationships), and improves exact match from 44\% to 76\% compared to the ablated version of our system. Our recurrent memory helps in particular for the choice selection task, improving from 71\% to 83\% accuracy.

We also compare the performance of our full and ablated systems on the tasks of resolving the partner's referring expressions and mention prediction, with results given in \Appendix{app:evaluation-other}.

\subsection{Evaluation in Self-Play}
\label{sec:results-selfplay}
\begin{table}
\begin{adjustbox}{width=0.95\columnwidth,center}
\begin{tabular}{@{}lccc@{}}
\toprule
Model & \#Shared=4 & \#Shared=5 & \#Shared=6 \\
\midrule
U\&A (2020) & 50.7$\pm$2.0 & 66.0$\pm$1.9 & 83.5$\pm$1.5\\
\midrule
F\nomem\nostruc & 42.3$\pm$2.1 & 57.0$\pm$2.1 & 75.4$\pm$1.1 \\
F\nomem& 52.6$\pm$1.5 & 67.1$\pm$1.9 & 84.1$\pm$1.6 \\
\modelfull & 58.5$\pm$2.7 & 71.6$\pm$2.9 & 86.8$\pm$1.8 \\
\modelprag & 62.4$\pm$2.2 & 74.7$\pm$2.7 & 90.9$\pm$1.4 \\
\midrule
Human & 65.8 & 77.0 & 87.0 \\
\bottomrule
\end{tabular}
\end{adjustbox}
\caption{\label{tbl:selfplay_results}
Task success rates in automatic self-play evaluations, by difficulty of context (the number of items shared in the players' views).
Our \modelfull model outperforms past work:
U\&A (2020) is our tuned reimplementation of \citet{udagawa2020annotated}.
Human shows success rates of trained human annotators in collecting the dataset \citep{udagawa2019natural}.
\vspace{-1em}
}
\end{table}

To evaluate systems on the full dialogue task, we first use
self-play, where a system is partnered with a copy of itself, following \citet{udagawa2020annotated}.
We evaluate systems on 3,000 world contexts, stratified into contexts with 4, 5, and 6 dots overlapping between the two agents' views, with 1,000 contexts in each stratification. 

Table \ref{tbl:selfplay_results} reports average task success (the fraction of times both agents chose the same dot at the end of the dialogue) averaged across the 10 copies of each model trained on the cross-validation splits. 
As in the corpus evaluation, we see substantial improvements to our system from the structured referent prediction and the recurrent reference memory. Our Full system, without pragmatic generation, improves over the system of \citet{udagawa2020annotated} from 51\% to 58\% in the hardest setting, with a further improvement to 62\% when adding our pragmatic generation procedure.

\subsection{Human Evaluation}
\label{sec:human_evaluation}
Finally, we perform human evaluation by comparing system performance when playing with workers from Amazon's Mechanical Turk (MTurk). 
To conduct evaluation, we used 100 world states from the \#Shared=4 partition, and collected 718 complete dialogues by randomly pairing worker with one of the following three: our best-performing model in self-play (\modelprag), the model from \citet{udagawa2020annotated}, or another worker.

In order to ensure higher quality dialogues, and following \citet{udagawa2019natural}, we filtered workers by qualifications, showed workers a game tutorial before playing, and prevented dots from being selected within the first minute of the game.
We paid workers \$0.30 per game, with a bonus of \$0.15 if the dialogue was successful.
See \Appendix{sec:dialogue-examples} for sample dialogues.

\begin{figure}
    \centering
    \small
    \begin{tikzpicture}
\begin{axis}[
    title={Dialogue Success Rates in Human Evaluations},
    xlabel={Success \%},
    xbar=5pt,
    symbolic y coords={{U\&A (2020)},{Full+Prag},{Human}},
    ytick=data,
    width=7cm,
    height=3cm,
    bar width=10pt,
    nodes near coords={
        \pgfmathprintnumber[precision=1,fixed zerofill]{\pgfplotspointmeta}
    },
    every node near coord/.append style={anchor=west, inner xsep=12.0pt},
    enlarge x limits={value=.23,upper},
    xmin=0,
    enlarge y limits=.3,
    xtick style = {draw=none},
    ytick style = {draw=none},
]

    \addplot[draw=none, fill=blue!50,
      error bars/.cd,
      x dir=both,
      x explicit,
      ] coordinates {
        (30.6,{U\&A (2020)}) +- (3.09,0)  %
        (46.0,{Full+Prag}) +- (3.14,0) %
        (45.0,{Human}) +- (3.19,0) %
    };
\end{axis}
\end{tikzpicture}
\caption{Success rates of systems on the full dialogue game task when paired with human partners. Error bars show standard errors. Our \modelprag{} system achieves a 50\% relative performance improvement over past work (U\&A 2020: \citealt{udagawa2020annotated}). 
    }
    \label{fig:results-human}
    \vspace*{-1.5em}
\end{figure}
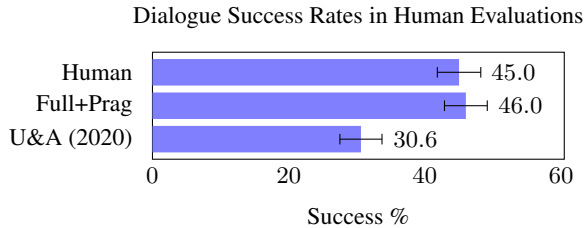

We compare systems based on the percentage of successful dialogues. 
The results, in \Figure{fig:results-human}, corroborate the trends observed in self-play. 
Both the models of U\&A (2020) and our \modelprag perform worse against humans than against agent partners in the automatic self-play evaluation, illustrating the importance of performing human evaluations. 
However, the trend is preserved, and we see that the \modelprag system substantially outperforms the U\&A (2020) model, resulting in a 50\% relative improvement in task success rate.
This difference is statistically significant at the $p\le0.05$ level using a one-tailed t-test.

\subsection{Success by Human Skill Level}
\label{sec:skill-analysis}
In \Section{sec:human_evaluation}, we compared our systems to a human population of MTurk workers. However, human populations themselves vary greatly based on many factors, including the day and time workers are recruited, training and feedback given to workers, and worker retention. 
One difference between our worker population and the population that produced the dataset is training. When collecting the dataset, \citet{udagawa2019natural} performed manual and individualized coaching of their MTurk workers which made them more effective at the game:
giving players personalized feedback on how to improve their game strategies, \eg ``please ask more clarification questions.''\footnote{Udagawa and Aizawa also manually removed around 1\% of dialogues where workers did not follow instructions. While we do not perform post-hoc manual filtering of the dialogues, in order to avoid introducing systematic bias that would favor or disfavor one of the systems we compare, an inspection of a subset of our collected dialogues indicates a similarly high fraction of our workers were making a good effort at the task.}
Manual coaching produced a high-quality corpus by increasing players' skill and obtained a success rate of 66\%; however coaching would make human evaluations difficult to replicate across works due to the labor, cost, and variability that coaching involves.

\begin{figure}[t]
  \centering

\includegraphics[width=1\linewidth]{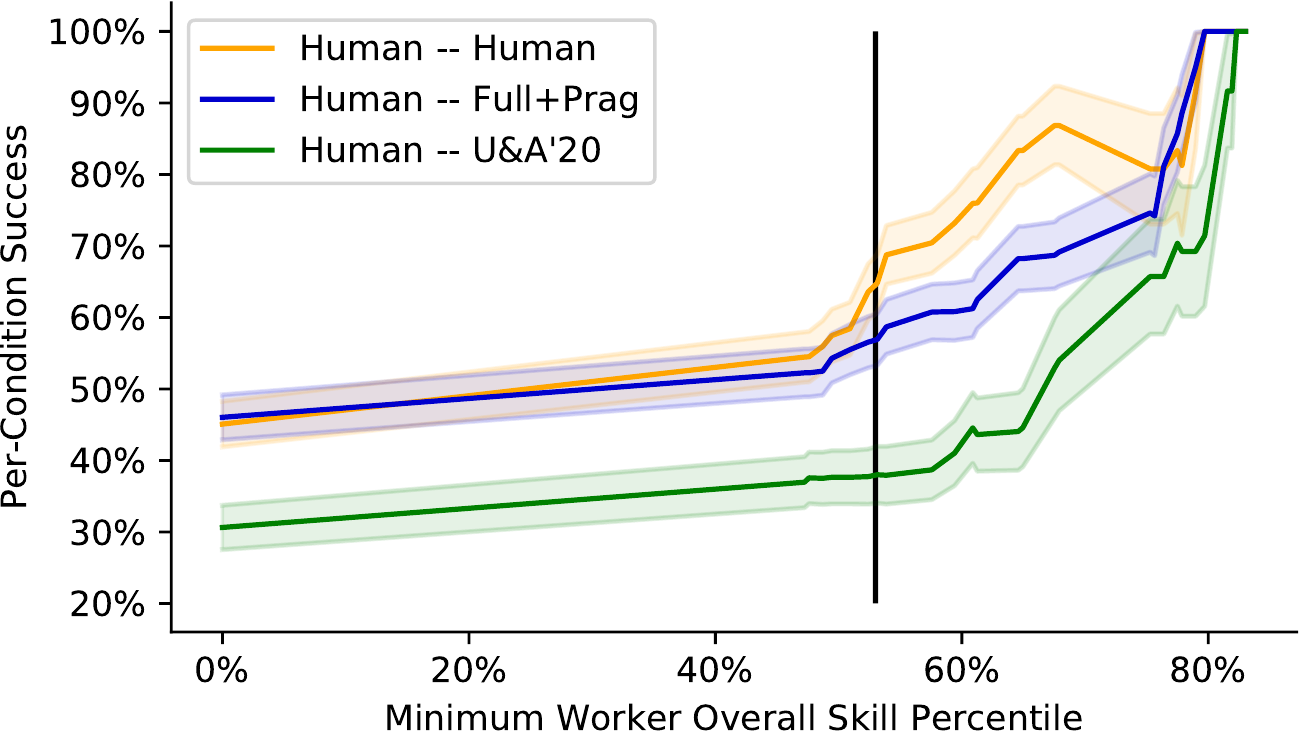}
\caption{\label{fig:human-filtering}
Success rates of human players against each system type, and other humans, with progressive filtering of humans by their overall success rate (across all conditions) along the x-axis. 
Shaded regions give standard errors.
 Our \modelprag{} system outperforms past work (U\&A 2020) at all levels.\footref{ft:filtering} %
\vspace*{-1.5em}
}
\end{figure}

In this section, we run a sweep of system comparisons of the form of \Section{sec:human_evaluation}, but on increasingly select sub-populations of MTurk workers.
Results are shown in Fig.~\ref{fig:human-filtering}. The x-axis gives the minimum skill percentile for a worker's games to be retained (with a worker's skill defined to be their average success across all games; see Appendix~\ref{sec:alternative-skill-analysis} for an alternative), so that the far left of the graph shows all workers (corresponding to the numbers in Fig.~\ref{fig:results-human}), the far right shows only those workers who won all of their games, and the black vertical line marks the player filtering needed to obtain a human-human success rate comparable to \citet{udagawa2019natural}. Our \modelprag{} system outperforms the model of \citet{udagawa2020annotated} at all player skill levels.\footnote{Until, by necessity, the point where filtering removes all workers who lost a game against any system. Differences between U\&A'20 and \modelprag{} are significant at the $p\le0.05$ level by a one-tailed t-test for minimum worker overall skills up to the 68th percentile.\label{ft:filtering}}
This result shows that, while more accomplished workers' overall success rates can be much higher than the success rate of our general worker population, in all cases the ordering between the two systems remained the same.

\section{Related Work}
\paragraph{Goal-oriented dialog.}

The modular approach that we use reflects the pipelined approach often used in goal-oriented dialogue systems \cite{young2013pomdp}.
Recent work on neural systems has also used structured and memory-based approaches \citep{bordes2016,he2018neg} including tracking entities identified in text \citep{williams2017hybrid,he2017mutualfriends}.
We also find improvements from an entity-centric approach with a structured memory, although our domain involves more challenging entity resolution and generation due to the spatial grounding.

\paragraph{Referring expressions.}
A long line of past work on referring expression grounding has tackled generation \cite{dale1989cooking,dale1995,viethen2011generating,krahmer-van-deemter-2012-computational}, interpretation \cite{schlangen2009incremental,liu2013modeling,kennington-schlangen-2015-simple} or both \cite{heeman-hirst-1995-collaborating,mao2016generation,yu2017joint}.
Closest to ours is the work of \citet{takmaz-etal-2020-refer}, which builds models for reference interpretation and generation in the rich PhotoBook corpus \cite{haber2019photobook}, focusing on a non-interactive setting with static evaluation on reference chains extracted from human-human dialogues.

\paragraph{Collaborative games.}
The closest work on dialogue systems for collaborative grounded tasks has focused on tasks with different properties from ours, as discussed in \Section{sec:introduction}.
A closely related task to the shared visual reference game we pursue here is the PhotoBook task \citep{haber2019photobook}, although a dialogue system has not been constructed for it.
Other work on grounded collaborative language games includes collection games \cite{Potts:2012,suhr-etal-2019-executing}, navigation and interactive question games \citep{thomason2019cvdn,nguyen2019hannah,ilinykh2019meet}, and construction tasks \citep{wang2017int,kim-etal-2019-codraw,narayan-chen-etal-2019-collaborative}.

\paragraph{Pragmatics.}
Our approach to pragmatics \citep{grice1975logic} builds on a large body of work in the RSA framework \citep{Frank12predictingpragmatic,GoodmanFrank2016-TICS}, which models how speakers and listeners reason about each other to communicate successfully. 
The most similar applications to ours in past work on computational pragmatics have been to single-turn grounded reference tasks (rather than dialogue), with much smaller and unstructured spaces of referents than ours,\footnote{Our setting has $2^7$ possible referents for each referring expression in the dialogue.} such as
discriminative image captioning \citep{v2017caption,andreas2016reasoning,cohn-gordon-etal-2018-pragmatically} and referent identification \citep{monroe-etal-2017-colors,mcdowell-goodman-2019-learning,white2020learning}.
Explicit speaker--listener models of pragmatics have also been used for dialogue, and while these approaches plan or infer across multiple turns (which our work does not do explicitly), they have either involved ungrounded settings \citep{kim-etal-2020-will} or constrained language \citep{vogel-etal-2013-emergence,khani2018pip}.

\section{Conclusion}
We presented a modular, reference-centric approach to a challenging partially-observable grounded collaborative dialogue task. 
Our approach is centered around a structured referent grounding module, 
which  
we use (1) to interpret a partner's utterances and (2) to enable a pragmatic generation procedure that encourages the agent's utterances to be able to be understood in context.
We perform, for the first time, human evaluations on the full dialogue task, finding that our system cooperates with people substantially more successfully than a system from past work and---in aggregate---achieves a success rate comparable to pairs of human partners.

While our results are encouraging, there is still much room for improving all systems in their interactions with people on this challenging task.
As the examples in \Appendix{sec:dialogue-examples} illustrate, people use sophisticated conversational strategies to build common ground \cite{clark1986referring,traum1994computational,clark1996using} when they interact with each other, producing utterances that play multiple conversational roles and performing complex reasoning.
To better plan utterances \cite{cohen1979elements} and more accurately infer the partner's state \cite{allen1980analyzing}, we suspect it will be helpful to extend the single-step pragmatic utterance planning and implicit inference procedures that we use here: planning over longer time horizons, performing more explicit reasoning under uncertainty, and learning richer models of the full range of speech acts that people use.
Future work might continue to explore these directions on this task and other similarly challenging tests of collaborative grounding.

\vspace{-0.3em}
\section*{Acknowledgments}
\vspace{-0.3em}

We are grateful to Takuma Udagawa for sharing information about the corpus collection, and to Yoav Artzi, Chris Potts, Aida Nematzadeh, the Berkeley NLP group, and the anonymous reviewers for helpful suggestions and feedback.
This work was supported by DARPA through the XAI program and by a Google PhD Fellowship to DF.

\bibliography{custom}
\bibliographystyle{acl_natbib}

\appendix

\clearpage
\section{Model Details}
\label{sec:appendix}
\label{app:model-details}
\subsection{Structured CRF}
\label{app:structured-crf}
\paragraph{Dot Configurations.}
Dot configuration potentials $\psi(r, z)$ are composed of two terms: $R(r,z)$ which decomposes into functions of pairwise relationships between the dots (whether active or not) in the context $w$ and the text features $z$, and $A(r,z)$ which is a function of all active dots in the referent: 
$$\psi(r,z) = R(r,z) + A(r,z)$$
The pairwise relationships are
\begin{align*}
  R(r,z) &= \sum_{i=1}^{N-1} \sum_{j=i+1}^N \alpha(r,z,i,j)
\end{align*}
where $N$ is the number of dots in view (7) and $\alpha$ is a scalar-valued neural function of the text features and whether the dots indexed by $i$ and $j$ are active in the referent $r$:
\begin{align*}
  \small
  \alpha(r,z,i,j) = \left\{\begin{array}{lr}
      p(z,i,j)_0, & r(i) \land r(j) \\
      p(z,i,j)_1, & \neg r(i) \land \neg r(j) \\
      p(z,i,j)_2, & \text{otherwise}
    \end{array}
    \right.
\end{align*}
$p$ is a 3-dimensional vector produced by an MLP:
\begin{align*}
  p(z,i,j) &= \mlp_\psi([w(i)-w(j), z])
\end{align*}
The active dot potential $A$ is designed to model group properties such as cardinality and common attributes, which other work has found useful on this and similar tasks \cite{Tenbrink03group-basedspatial,udagawa2020spatial}. We define the potential as
\begin{align*}
  A(r,z) &= \mlp_A([w(r), e(r)])
\end{align*}
where $w(r)$ is the mean of the feature values for the active dots in $r$, $\frac{1}{|r_{active}|} \sum_{d \in r_{active}} w(d)$ and $e(r)$ is a learned 40-dimensional embedding for the discrete count of active dots in $r$.
\paragraph{Configuration Transitions.}
The configuration transition potential $\omega(r^{k:k+1}, z^{k:k+1})$ is similar to the dot configuration potential above but bridges the dots in referents $k$ and $k+1$. It is the sum of two terms: $\omega(r^{k:k+1}, z^{k:k+1}) = S(r^{k:k+1}, z^{k:k+1}) + B(r^{k:k+1}, z^{k:k+1})$. First is $S$, which decomposes into pairwise relationships between dots across referents $r^k$ and $r^{k+1}$:
\begin{align*}
  S(r^{k:k+1}, z^{k:k+1}) = \sum_{i=1}^N \sum_{j=1}^N \beta(r^{k:k+1}, z^{k:k+1}, i, j) 
\end{align*}

\begin{align*}
  \small
  \beta(r^{k:k+1},z^{k:k+1},i,j) = \left\{\begin{array}{lr}
      q_0, & r^k(i) \land r^{k+1}(j) \\
      q_1, & \neg r^k(i) \land \neg r^{k+1}(j) \\
      q_2, & \text{otherwise}
    \end{array}
    \right.
\end{align*}
$q$ (short for $q(z^{k:k+1},i,j)$) is, like $p$ in Dot Configurations, a 3-dimensional vector produced by an MLP:
\begin{align*}
  q &= \mlp_\omega([w(i) - w(j), z^{k}-z^{k+1}])
\end{align*}
Next is $B$, which is a function of the feature centroids of the active dots in referents $k$ and $k+1$. $B$ is defined analogously to $A$ in Dot Configurations:
\begin{align*}
\small
  B(&r^{k:k+1}, z^{k:k+1}) \\
  &= \mlp_B([w(r^k) - w(r^{k+1}), z^{k}-z^{k+1}])
\end{align*}
with $w(r)$ again giving the mean of the feature values for the active dots in $r$.
We fix $B(r^{k:k+1}, z^{k:k+1}) = 0$ if $|r_{active}^k| > 3$ or $|r_{active}^{k+1}| > 3$, which had little effect on model accuracy but improves memory efficiency as it substantially reduces the number of group-pairwise relationships that need to be computed.

\paragraph{Inference.}
We compute the normalizing constant for the CRF distribution by enumerating the possible $2^7$ assignments to each $\zdotconfig^k$ to compute the $\phi$, $\psi$, and $\omega$ potential terms, which can be performed efficiently on a GPU. To compute the normalizing constant, which sums over all combinations of assignments to these $\zdotconfig^k$, we use the standard linear-chain dynamic program. In training, we backpropagate through the enumeration and dynamic program steps to pass gradients to the parameters of the potential functions. 

\subsection{Referent Memory}
\label{app:referent-memory}
The function $\iota$ collapses predicted values for the dot $d$ over $K$ referents into a single representation for the dot, which we do in two ways: by max- and average- pooling predicted values for $d$ across the $K$ referents. We also obtain the prediction values in two ways: by taking the argmax structured prediction from $\zlistener$, and by taking the argmax predictions from each dot's marginal distribution. We found that using these ``hard'' argmax predicted values gave slightly better results in early experiments than using the ``soft'' probabilities from \zlistener. 
In combination, these give four feature values as the output of $\iota(d, \zdot_t)$.

\subsection{Utterance Generation Module}
\label{app:utterance-generation}
We first use a bidirectional LSTM \cite{hochreiter1997lstm} to encode the sequence of $K$ referents-to-mention $\zdot_{t+1}=r_{t+1}^{1:K}$, using the inputs at each position $k \in [1, K]$ a mean-pooled representation of the world context embeddings for the active dots in the referent: $\frac{1}{|r_{t+1}^k|} \sum_{d \in r_{t+1}^k} w(d)$,
to produce a sequence of encoded vectors $y_{t}^{1:K}$. 
We make gated updates to the decoder's initial state, updating it with (i) a linear projection of the forward and backward vectors for $y_{t}^1$ and $y_{t}^K$, representing the referent context and (ii) an embedding for the discrete confirmation variable $\zcon_{t+1}$.

\subsection{Implementation Choices}
For our reimplementation of the system of \citet{udagawa2020annotated} in a shared codebase with our system, we replace their $tanh$ non-linearities with \texttt{ReLU}s and use PyTorch's default initializations for all parameters.
These improve performance across all evaluation conditions in comparison to the reported results. 

For our system, we use separate word-level recurrent models, a Reader and a Writer, to summarize the dialogue history. The Reader is bidirectional over each utterance, and is used in the reference resolution and choice selection modules. The Writer is unidirectional, and is used in the mention selection and utterance generation modules.

\subsection{Hyperparameters}
\label{app:hyperparameters}
\begin{table}[h]
  \small
  \centering
  \begin{tabular}{lc}
    \toprule
    \multicolumn{2}{l}{\bf Recurrent Unit Hyperparameters} \\
    Reader GRU size & 512 \\
    Writer GRU size & 512 \\
    Mention decoder $\text{RNN}_M$ size & 512 \\
    Referent memory $\text{RNN}_C$ size & 64 \\
    Confirmation embedding $c$ size & 512 \\
    \midrule
    \multicolumn{2}{l}{\bf CRF Hyperparameters} \\
    $\text{MLP}_\phi$ hidden layers & 2 \\
    $\text{MLP}_\phi$ hidden size & 256 \\
    $\text{MLP}_\phi$ dropout & 0.5 \\
    $\text{MLP}_\psi$ and $\mlp_\omega$ hidden size & 64 \\
    $\text{MLP}_\psi$ and $\mlp_\omega$ dropout & 0.2 \\
    $\text{MLP}_\psi$ and $\mlp_\omega$ hidden layers & 1 \\
    $\text{MLP}_A$ and $\mlp_B$ hidden size & 64 \\
    $\text{MLP}_A$ and $\mlp_B$ dropout & 0.2 \\
    $\text{MLP}_A$ and $\mlp_B$ hidden layers & 1 \\
    \midrule
    \multicolumn{2}{l}{\bf Generation Hyperparameters} \\
    Sampling temperature in $\zspeaker$  & 0.25 \\
    \# Utterance candidates, $N_\zutt$ & 100 \\
    \# Referent candidates, $N_r$ & 20 \\
    Mention weight, $w_M$ & 0 \\
    Speaker weight, $w_S$ & $1\times10^{-3}$ \\
    Listener weight, $w_L$ & $1 - w_S$ \\
    Early-stopping threshold, $\tau$ & 0.8 \\
    \bottomrule
\end{tabular}
\vspace*{-1em}
\end{table}
\subsection{Training Details}
\label{app:training-details}

For our full system and ablations, we train on each cross-validation fold for 12 epochs using the Adam optimizer \citep{kingma2014adam} with an initial learning rate of $1\times 10^{-3}$ and early stopping on the fold's validation set. 
Our loss function is a weighted combination of losses for the subtask objectives: 
\par\nobreak
\vspace{-1em}
{\small
\begin{align*}
\mathcal{L} &= w_S \log P_S(s | ...) + \\
&\frac{1}{T} \sum_{t=1}^T (\log P_R(r_t | ...) + \log P_M(r_t | ...) + \log P_U(u_t | ...) ),
\end{align*}
}
where $w_S$ is a hyperparameter which we set to $\frac{1}{32}$ following \citet{udagawa2020annotated} and we have omitted conditioning contexts from the probability distributions for brevity; see \Section{sec:decomposing-turns} for the full contexts.
We decay the learning rate when the loss plateaus on validation data.

We train models on a Quadro RTX 6000 GPU. Training takes around 1 day for models that use the structured CRF, and several hours without the structured CRF. Self-play evaluation takes around 1 hour.

\section{Evaluation for Other Subtasks}
\label{app:evaluation-other}
\begin{table}
  \small
\begin{adjustbox}{width=\columnwidth,center}
\ra{1.1}
\begin{tabular}{@{}lccc@{}} 
\toprule
& \multicolumn{2}{c}{Partner Refs.} & Next Refs \\
Model  &  Acc. & Ex. & Ex. \\
\midrule
F\nomem\nostruc& 87.3$\pm$0.3 & 41.8$\pm$0.9 & \phantom{0}4.8$\pm$1.0 \\
F\nomem& 90.6$\pm$0.3 & 65.2$\pm$1.0 & 23.5$\pm$2.0\\
\modelfull & 91.2$\pm$0.4 & 67.0$\pm$1.0 & 31.1$\pm$1.0\\
\bottomrule
\end{tabular}
\end{adjustbox}
\caption{\label{tbl:static_results2}
Accuracies for resolving referents in the \emph{partner}'s view (dot-level accuracy Acc. and exact match Ex.) and predicting the next referents to mention in the dialogue (Next Refs Ex.) in 10-fold cross-validation on the corpus of human--human dialogues. 
Our \modelfull benefits from its recurrent referent memory (outperforming F\nomem) and structured referent prediction module (outperforming F\nomem\nostruc).
}
\end{table}

Table \ref{tbl:static_results2} gives performance accuracies for resolving referents in the \emph{partner}'s view (dot-level accuracy Acc. and exact match Ex.) and predicting the next referents to mention in the dialogue (Next Refs Ex.) in 10-fold cross-validation on the corpus of human--human dialogues. 
We observe improvements from both the recurrent memory (comparing F-Mem to Full) and the structured referent prediction module (comparing F-Mem-Struc to Full) on both tasks.

\section{Pragmatic Generation}
\label{app:pragmatic-generation}
We give pseudocode for the pragmatic generation procedure (\Section{sec:pragmatics}) in \Algorithm{alg:praggeneration}. \Figure{fig:prag-gen} shows an example, showing 2 referents $\zdot$ (inputs to the \texttt{realize}) function on the left, and 3 utterances $\zutt$ sampled for each referent on the right.
Fewer than $N_r$ referent candidates may be evaluated (as in \Figure{fig:prag-gen}) if one $(\zdot, \zutt)$ pair is found with $L(\zdot, \zutt) \ge \tau$.

\SetAlCapSkip{0.6em}
\providecommand\algf{}
\begin{algorithm}[ht!]
\small
  \DontPrintSemicolon
  \SetKwFunction{generate}{generate}
  \SetKwFunction{realize}{realize}
  \SetKwProg{Function}{function}{:}{}
  \SetKwIF{If}{ElseIf}{Else}{if}{:}{else if}{else}{end if}
  \SetKw{KwTo}{in}\SetKwFor{For}{for}{\string:}{}%
  \SetKwRepeat{Do}{do}{while}
  \text{hyperparameters: $N_r$, $N_u$, $\tau$}\;
  
  \Function{\generate{$\zmem, \zutt_{1:t-1}, \zcon_t, \zwor$}}{
  $(\hat \zdot, \hat \zutt, \hat s) \gets (\text{None}, \text{None}, -\infty)$\;
  \For{$\zdot \in \topk_{N_r} \zmention(\zdot | \zutt_{1:t-1}, \zmem, \zcon, \zwor)$}{
  $\zutt, s \gets $\realize{$\zdot$}\;
  \If{$s > \hat s$}{
  $(\hat \zdot, \hat \zutt, \hat s) \gets (\zdot, \zutt, s)$\;
          \If{$s \ge \tau$}{
            {\algf \textbf{break}}\;
          }
      }
  }
  \KwRet $\hat \zdot, \hat \zutt$
  }
  
  \Function{\realize{$\zdot$}}{
  \For{$k \in 1\ldots N_u$} {
  $\zutt^{(k)} \thicksim \zspeaker(\cdot \mid \zdot)$
  }
  $\hat \zutt \gets \argmax_{\zutt^{(k)}} L(\zdot, \zutt^{(k)})$ (Eqn.~\ref{eqn:pragmatic-objective} in Sec.~\ref{sec:pragmatics})\;
  $\hat s \gets \max_{\zutt^{(k)}} L(\zdot, \zutt^{(k)})$\;
  \KwRet $(\hat \zutt, \hat s)$ \;
  }
  \caption{\label{alg:praggeneration} Our pragmatic generation procedure chooses a sequence of referents $\zdot$ to describe, and an utterance $\zutt$ to describe them, to optimize the objective $L(\zdot, \zutt)$ (\Equation{eqn:pragmatic-objective} in \Section{sec:pragmatics}) using candidates from the models $\zmention$ and $\zspeaker$ and an early stopping search with threshold $\tau$.
}
\end{algorithm}

We used self-play evaluation on one of the cross-validation splits to tune the early-stopping threshold $\tau$, selecting from among the values \{0.0, 0.6, 0.7, 0.8, 0.9\}. The optimal value was $\tau=0.8$, but the success rate in self-play was fairly robust to the value chosen (including $\tau=0.0$, which results in performing pragmatic search only over those utterances for the single highest-scoring referent sequence under $P_M$), with a range of about 2\%. We did not evaluate without early-stopping (searching over all candidate reference sequences and utterances) as this would have made generation too computationally expensive to be feasible in both self-play and human evaluations. 

\section{Alternative Skill Analysis}
\label{sec:alternative-skill-analysis}

\begin{figure}[ht]
  \centering

\includegraphics[width=1\linewidth]{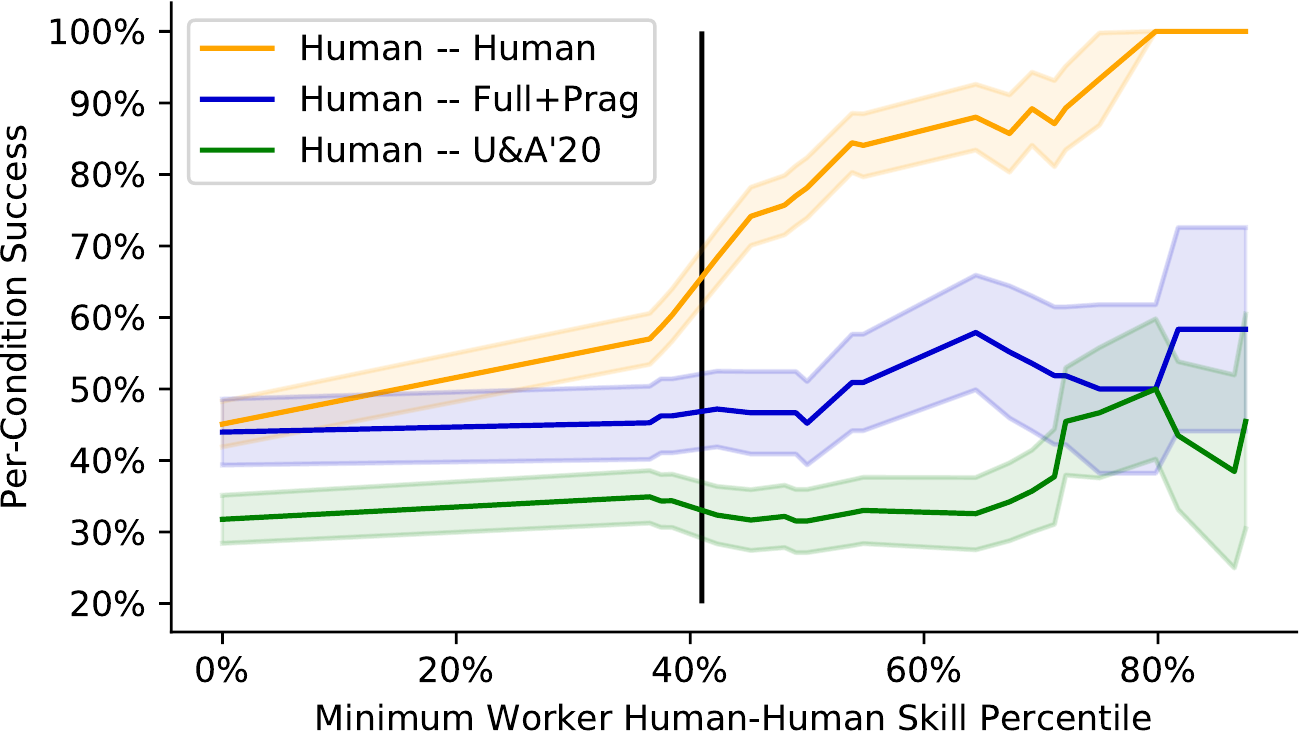}
\caption{\label{fig:alternative-human-filtering}
Success rates of human players against each system type, and other humans, with progressive filtering of humans by their overall success rate (when partnered with other humans) along the x-axis. 
Shaded regions give standard errors.
 Our \modelprag{} system outperforms the model from \citet{udagawa2020annotated} at all levels.\footref{ft:filtering} %
}
\end{figure}

In \Section{sec:skill-analysis}, we compared systems on increasingly select sub-populations of MTurk workers, selected by their average success across all conditions (whether playing with other humans or one of the two system types). 
In this section, we run a similar analysis but select workers by their average success when paired with \emph{human partners only}. 
Results are shown in \Figure{fig:alternative-human-filtering}. The x-axis gives the minimum skill percentile for a worker's games to be retained, with skill defined by a worker's average success when paired \emph{with other human workers}. The far left of the graph shows all workers,\footnote{After filtering to remove any workers who did not play at least one game with another human worker.} the far right shows only those workers who won all of their games when paired with other workers, and the black vertical line marks the player filtering needed to obtain a human-human success rate comparable to \citet{udagawa2019natural}. As we saw in \Section{sec:skill-analysis}, our \modelprag{} system outperforms the model of \citet{udagawa2020annotated} at all worker skill levels.
However, focusing on the sub-population of workers who are successful when paired with other humans (the right side of \Figure{fig:alternative-human-filtering}) reveals a gap between humans and our system: humans who are successful when partnering with other humans are substantially less successful when partnering with our \modelprag{} system (and even less successful when partnering with the model of U\&A'20).
This indicates room for improvement on the task, as we want to build a system that can collaborate as well as humans with any population of human partners.

\section{Dialogue Examples}
\label{sec:dialogue-examples}
\begin{figure*}[h]
\centering
\begin{subfigure}[t]{0.48\textwidth}
  \centering
\begin{tikzpicture}
\node[inner sep=0pt] (agent_0) at (0,0)
{\includegraphics[width=0.95\columnwidth,trim=0 0 0 0, clip]{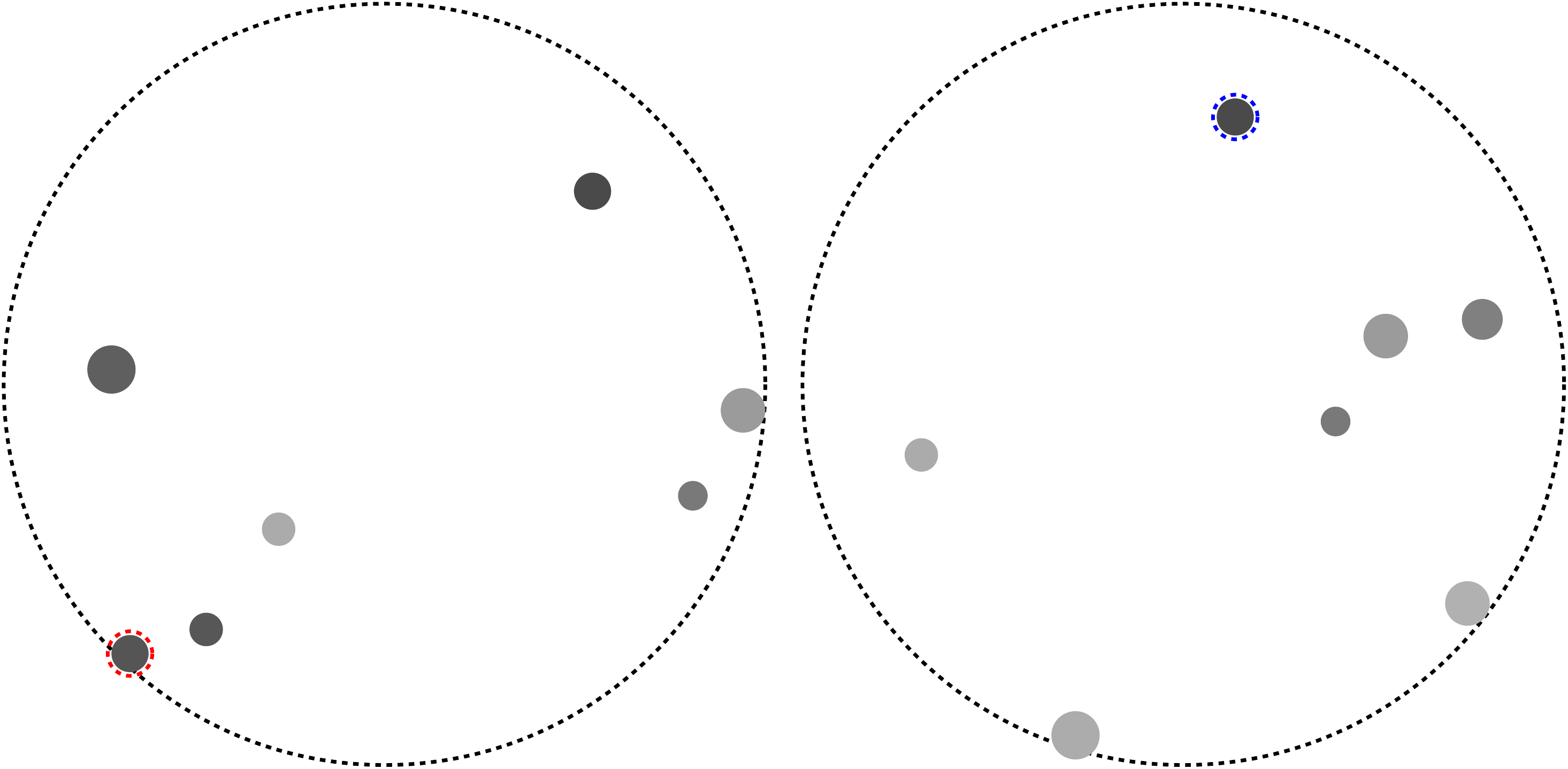}};
\node [below,color=red] at (-1.5,-2) {Human's view};
\node [below,color=blue] at (1.5,-2) {U\&A20's view};
\end{tikzpicture}
\small
\begin{tabular}{@{}l@{}l@{}}
\toprule
{\color{blue} U\&A'20}:\;& i have a large light gray dot with a \\& small black dot above it and to the left \\
{\color{red} Human}:\;& I don't see that.  I have a dark gray dot \\& on the line between 7 and 8 o'clock. \\
{\color{blue} U\&A'20}:\;& yes \\
{\color{red} Human}:\;& Let's pick that! \\
{\color{blue} U\&A'20}:\;& ok \\
{\color{red} Human}:\;&SELECT {\color{red} red} \\
{\color{blue} U\&A'20}:\;&SELECT {\color{blue} blue} \\
\bottomrule
\end{tabular}
\caption{
\label{fig:ua-failure}
An unsuccessful dialogue between a human and the system of \citet{udagawa2020annotated}.
}
\end{subfigure}
\hspace{3pt}
\begin{subfigure}[t]{0.48\textwidth}
  \centering
\begin{tikzpicture}
\node[inner sep=0pt] (agent_0) at (0,0)
{\includegraphics[width=0.95\columnwidth,trim=0 0 0 0, clip]{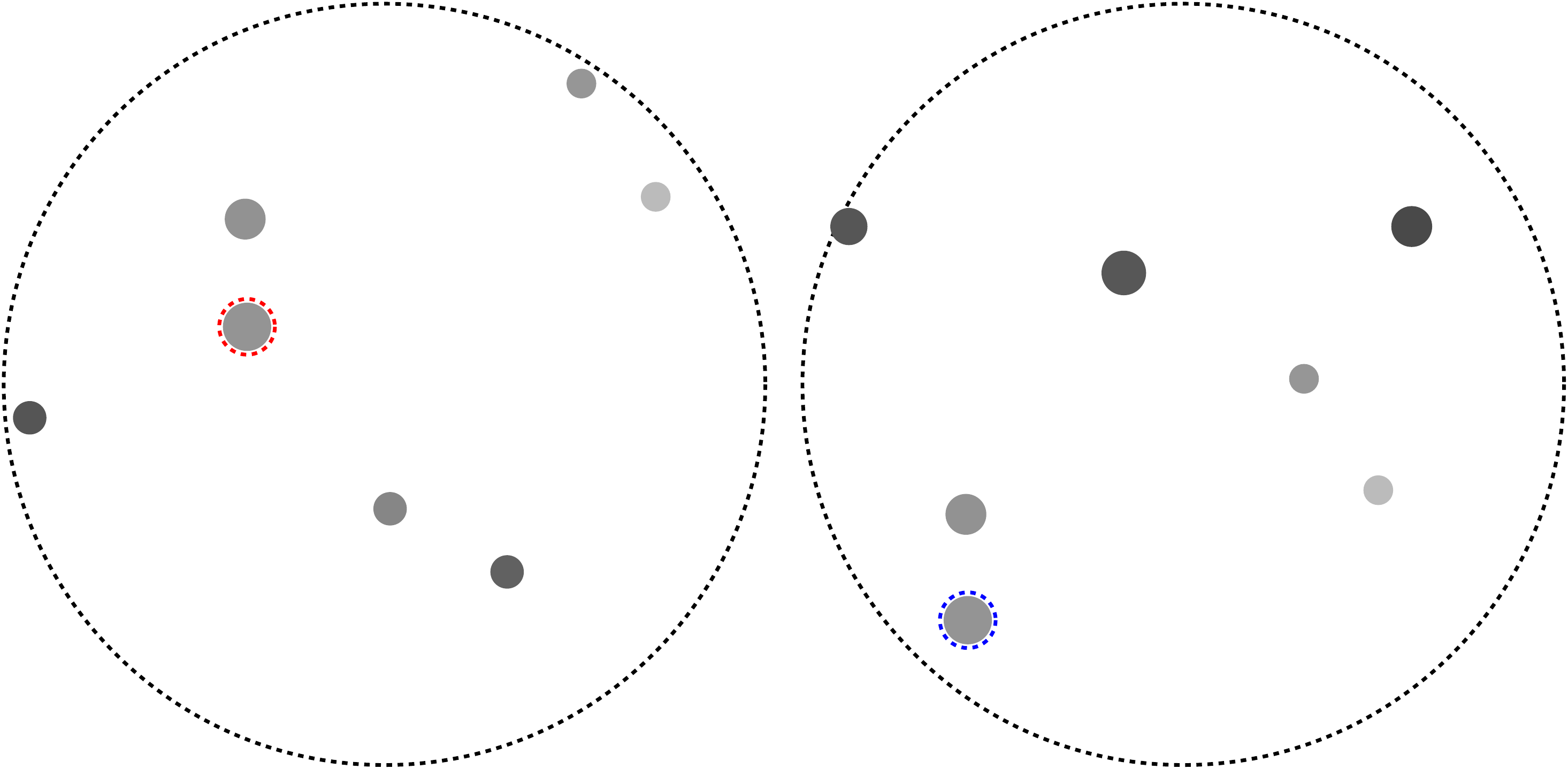}};
\node [below,color=red] at (-1.5,-2) {Human's view};
\node [below,color=blue] at (1.5,-2) {Full+Prag's view};
\end{tikzpicture}
\small
\begin{tabular}{@{}l@{}l@{}}
\toprule
{\color{blue} U\&A'20}:\;& i have a large black dot with a smaller \\& light grey dot to the left of it \\
{\color{red} Human}:\;& I don't see that.  I have a medium dark \\& dor to the left of all the others. \\
{\color{blue} U\&A'20}:\;& i have a large black dot with a smaller \\& one that is a little lighter and to the left of it . \\
{\color{red} Human}:\;& I have a large light gray with a medium \\& light gray above it. \\
{\color{blue} U\&A'20}:\;& yes , let 's click the larger one \\
{\color{red} Human}:\;&SELECT {\color{red} red} \\
{\color{blue} U\&A'20}:\;&SELECT {\color{blue} blue} \\
\bottomrule
\end{tabular}
\caption{
\label{fig:ua-success}
A successful dialogue between a human and the system of \citet{udagawa2020annotated}.
}

\end{subfigure}
\begin{subfigure}[t]{0.48\textwidth}
  \centering
\begin{tikzpicture}
\node[inner sep=0pt] (agent_0) at (0,0)
{\includegraphics[width=0.95\columnwidth,trim=0 0 0 0, clip]{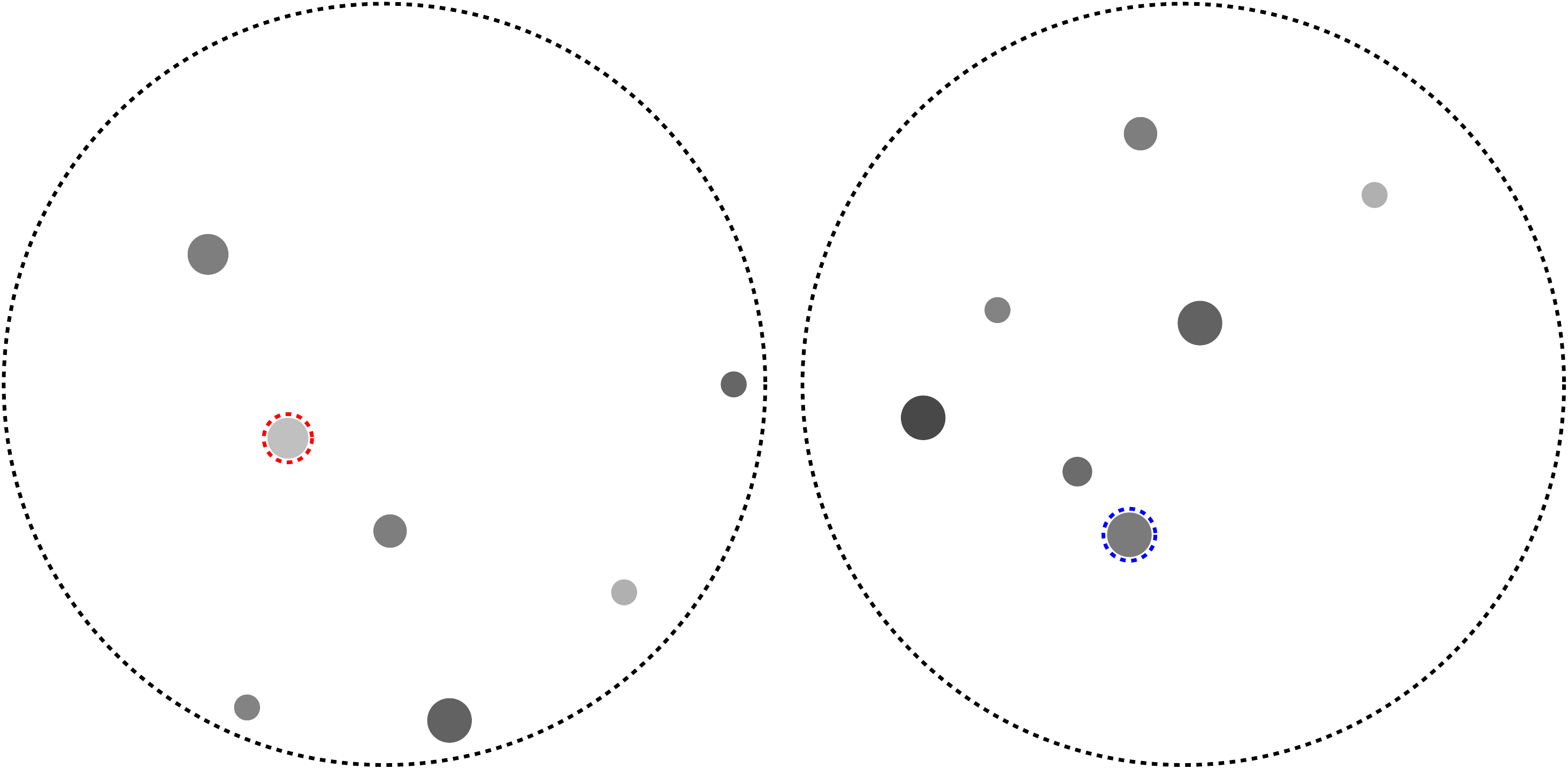}};
\node [below,color=red] at (-1.5,-2) {Human's view};
\node [below,color=blue] at (1.5,-2) {Full+Prag's view};
\end{tikzpicture}
\small
\begin{tabular}{@{}l@{}l@{}}
\toprule
{\color{blue} \modelprag}:\;& large black dot with a small light \\ & grey dot above it \\
{\color{red} Human}:\;& Also have a large gray dot with 3 \\ & smaller dots around it \\
{\color{blue} \modelprag}:\;& i do n't have that \\
{\color{red} Human}:\;& Medium sized and light gray dot \\ & towards the middle  \\
{\color{blue} \modelprag}:\;& yes , let 's pick the larger one \\
{\color{red} Human}:\;&SELECT {\color{red} red} \\
{\color{blue} \modelprag}:\;&SELECT {\color{blue} blue} \\
\bottomrule
\end{tabular}
\caption{
\label{fig:ours-failure}
An unsuccessful dialogue between a human and our \modelprag system.
}
\end{subfigure}
\hspace{3pt}
\begin{subfigure}[t]{0.48\textwidth}
  \centering
\begin{tikzpicture}
\node[inner sep=0pt] (agent_0) at (0,0)
{\includegraphics[width=0.95\columnwidth,trim=0 0 0 0, clip]{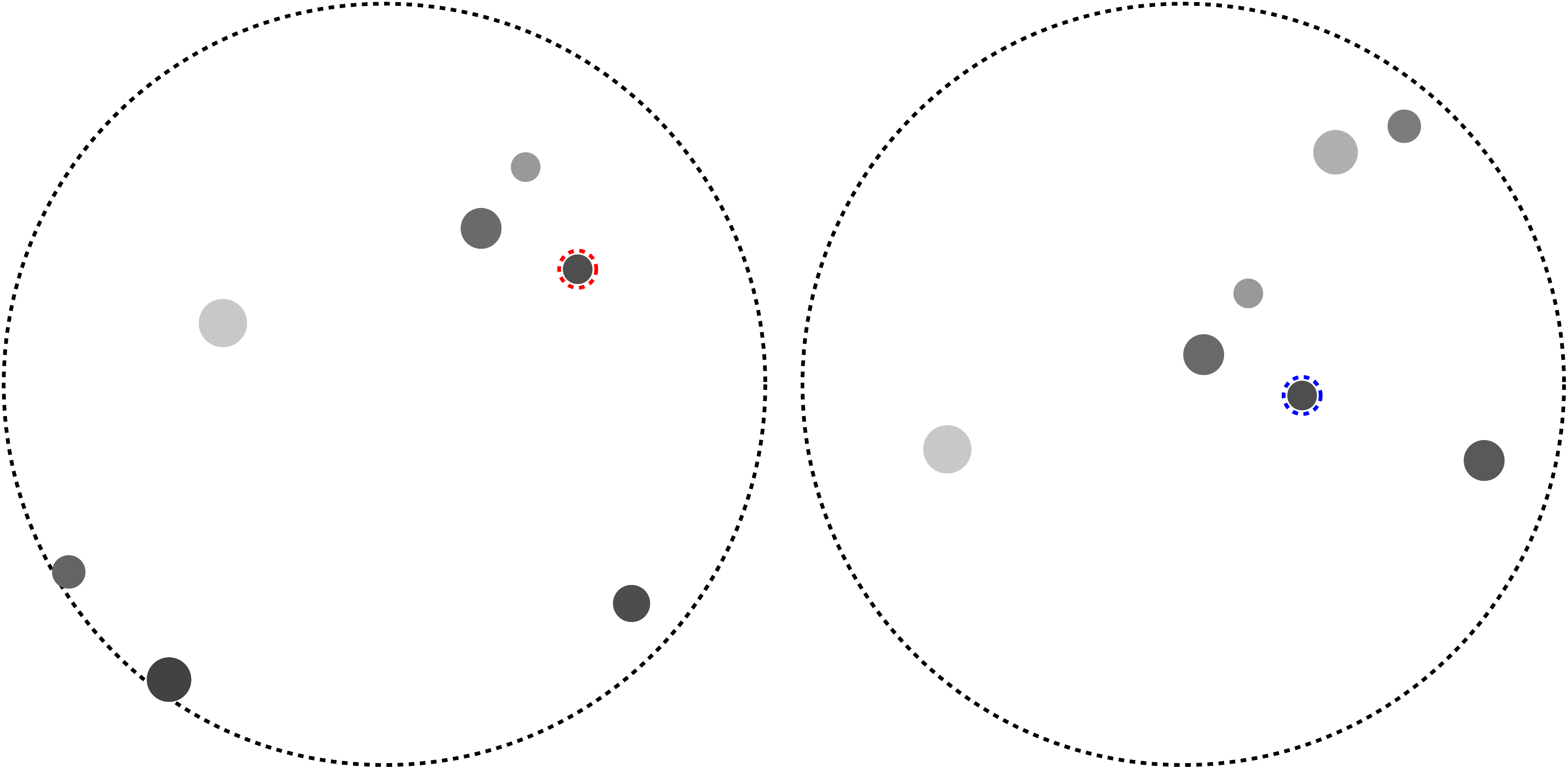}};
\node [below,color=red] at (-1.5,-2) {Human's view};
\node [below,color=blue] at (1.5,-2) {Full+Prag's view};
\end{tikzpicture}
\small
\begin{tabular}{@{}l@{}l@{}}
\toprule
{\color{blue} \modelprag}:\;& i have a group of 3 dots , a tiny \\ &light grey and a large dark grey \\
{\color{red} Human}:\;& yes me too \\
{\color{blue} \modelprag}:\;& let us select the small black \\
{\color{red} Human}:\;& okay \\
{\color{blue} \modelprag}:\;& pick the small black dot \\
{\color{red} Human}:\;& okay i will \\
{\color{blue} \modelprag}:\;& ok \\
{\color{red} Human}:\;&SELECT {\color{red} red} \\
{\color{blue} \modelprag}:\;&SELECT {\color{blue} blue} \\
\bottomrule
\end{tabular}
\caption{
\label{fig:ours-success}
A successful dialogue between a human and our \modelprag system.
}
\end{subfigure}
\caption{
\label{fig:system-examples} 
Example dialogues collected during our human evaluation (\Section{sec:human_evaluation}) of the dialogue systems. We show one unsuccessful (left) and one successful (right) example for each system. The top row is our reimplementation of \citet{udagawa2020annotated} and the bottom row is \modelprag, our full system with pragmatic inference.}
\end{figure*}

\begin{figure*}
\begin{subfigure}[t]{0.48\textwidth}
  \centering
\begin{tikzpicture}
\node[inner sep=0pt] (agent_0) at (0,0)
{\includegraphics[width=0.95\columnwidth,trim=0 0 0 0, clip]{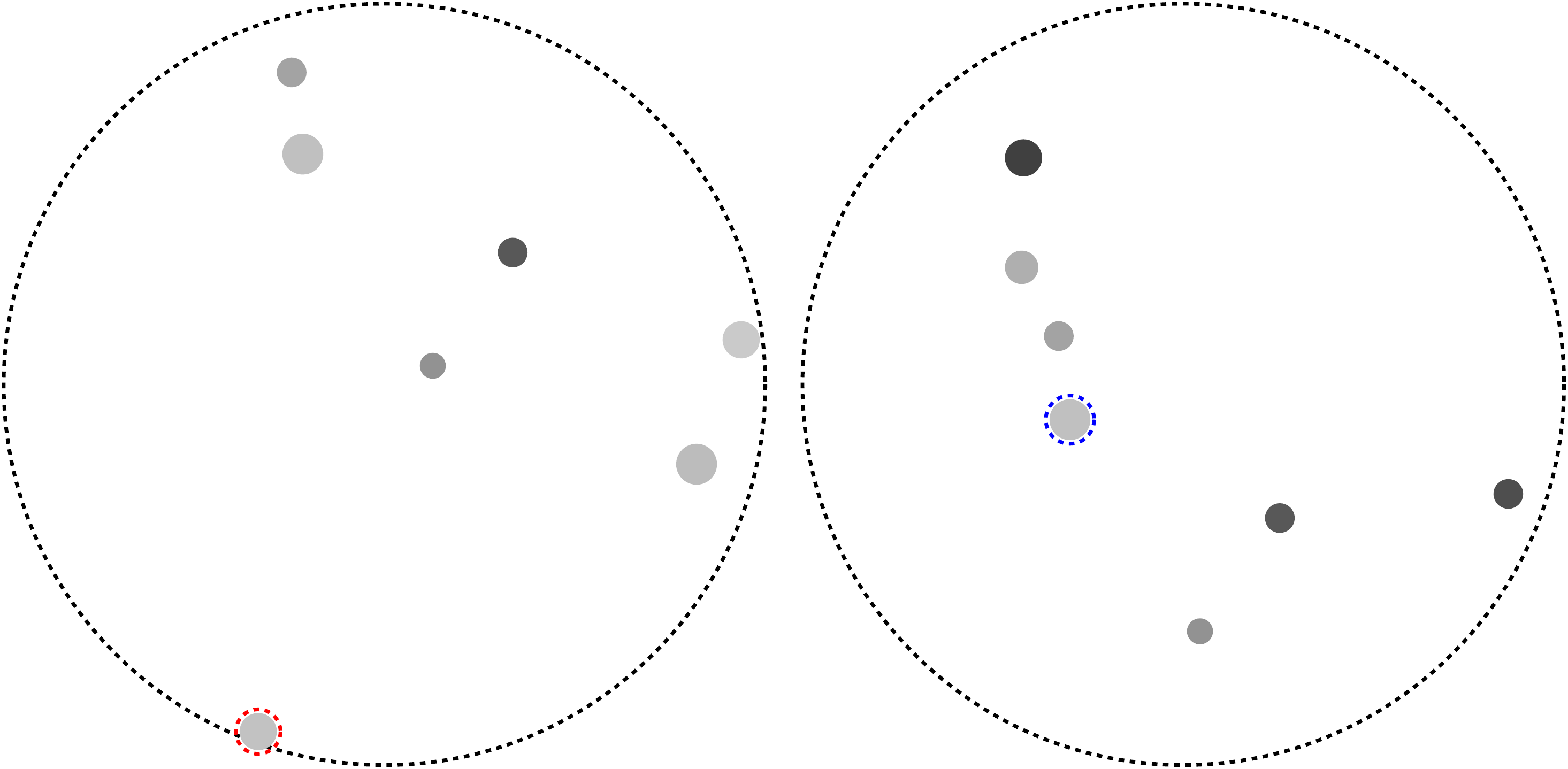}};
\node [below,color=red] at (-1.5,-2) {Human A's view};
\node [below,color=blue] at (1.5,-2) {Human B's view};
\end{tikzpicture}
\small
\begin{tabular}{@{}l@{}l@{}}
\toprule
{\color{blue} Human B}:\;& three light grey dots in a diagonal line \\
{\color{red} Human A}:\;& i dont have that but i have a black \\ & dot neer the top to the right, the only \\ & black dot in the circle \\
{\color{blue} Human B}:\;& i have two black dots. find something else \\
{\color{red} Human A}:\;& ok i have a light grey dot by itself \\ & at the bottom to the left. right on the line \\
{\color{blue} Human B}:\;& how big is it \\
{\color{red} Human A}:\;& its one of the bigger ones \\
{\color{blue} Human B}:\;& okay just pick it then \\
{\color{red} Human A}:\;& ok \\
{\color{blue} Human B}:\;&SELECT {\color{blue} blue} \\
{\color{red} Human A}:\;&SELECT {\color{red} red} \\
\bottomrule
\end{tabular}
\end{subfigure}
\hspace{3pt}
\begin{subfigure}[t]{0.48\textwidth}
  \centering
\begin{tikzpicture}
\node[inner sep=0pt] (agent_0) at (0,0)
{\includegraphics[width=0.95\columnwidth,trim=0 0 0 0, clip]{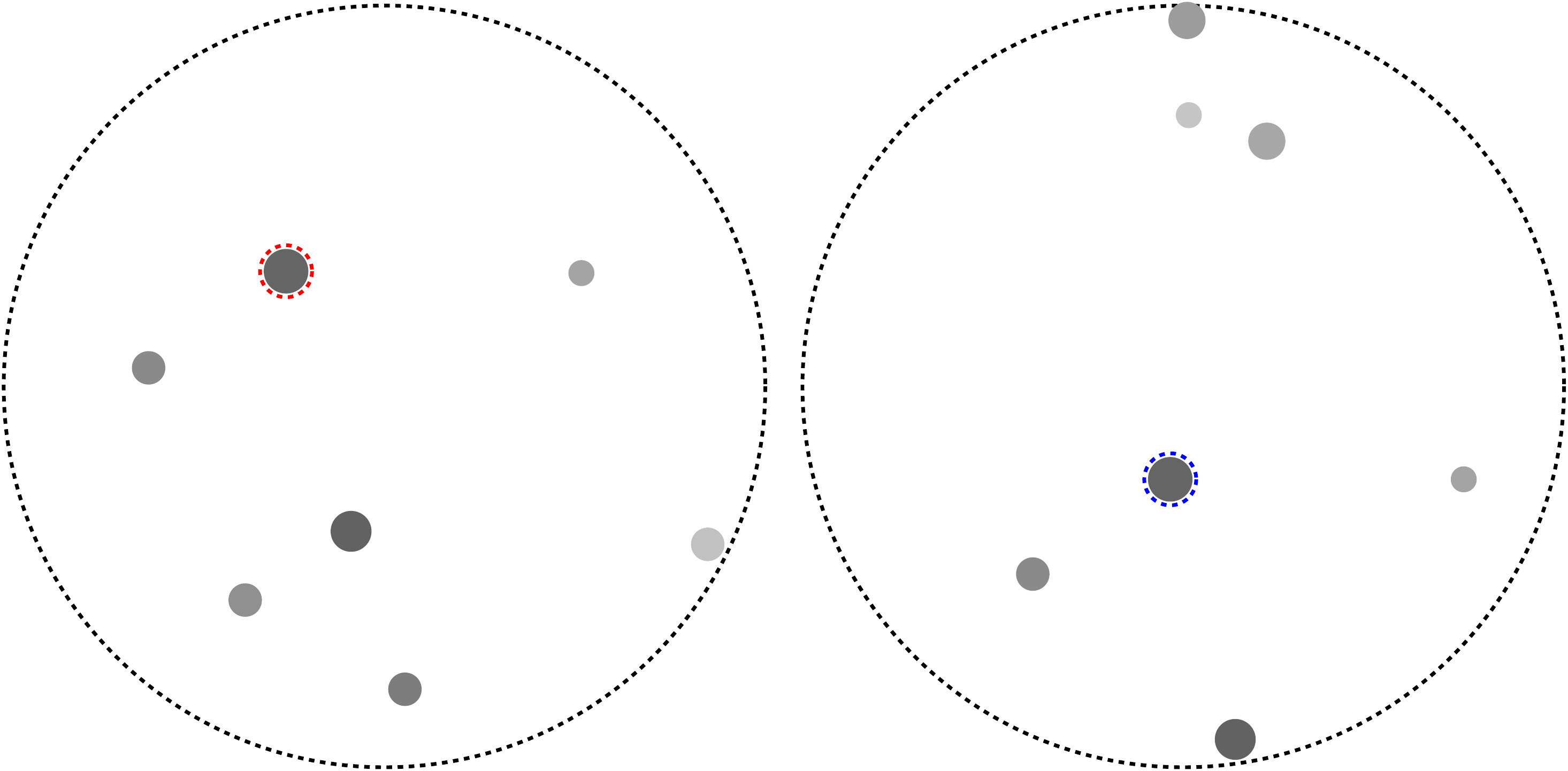}};
\node [below,color=red] at (-1.5,-2) {Human A's view};
\node [below,color=blue] at (1.5,-2) {Human B's view};
\end{tikzpicture}
\small
\begin{tabular}{@{}l@{}l@{}}
\toprule
{\color{blue} Human B}:\;& the smallest lightest grey dot.   it's \\& near a larger grey dot \\
{\color{red} Human A}:\;& did  you see 3 dark dots in same line \\
{\color{blue} Human B}:\;& no \\
{\color{red} Human A}:\;& did you see two larger black dots \\
{\color{blue} Human B}:\;& is one on top of the other? if   so, pick \\& the top one \\
{\color{red} Human A}:\;& ok \\
{\color{blue} Human B}:\;&SELECT {\color{blue} blue} \\
{\color{red} Human A}:\;&SELECT {\color{red} red} \\
\bottomrule
\end{tabular}
\end{subfigure}
\caption{Examples of unsuccessful (left) and successful (right) dialogues collected between pairs of people during our human evaluation (\Section{sec:human_evaluation}).}
\label{fig:human-examples} 
\end{figure*}
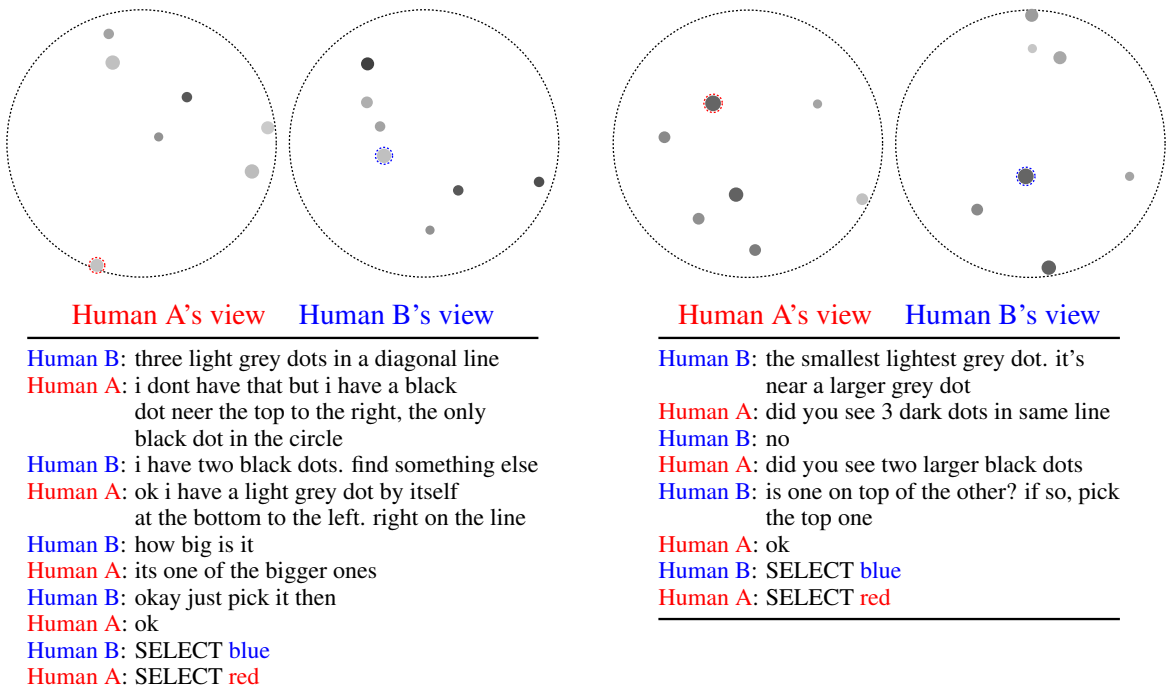

We show one successful and one failed dialogue from our human evaluations (\Section{sec:human_evaluation}) for each system (\Figure{fig:system-examples}) and from human--human pairs (\Figure{fig:human-examples}). 

As seen in these examples, descriptions from the baseline system (Figures \ref{fig:ua-failure} and \ref{fig:ua-success}) typically have a consistent syntactic structure (\eg ``i have a <size> <color> dot with a <size> <color> dot <spatial relation>'') but often do not correspond to the visual context. We suspect that it is difficult for this end-to-end generation model to simultaneously learn which dots to talk about (content selection) and how to describe them (surface realization) with the amount of training data available. 
Our \modelprag{} system (Figures \ref{fig:ours-failure} and \ref{fig:ours-success}) produces broader and generally more accurate utterances, which we attribute to our factored and pragmatic generation procedure.

Our system's utterances still have substantial qualitative differences from those in human-human dialogues (\Figure{fig:human-examples}), which---due to the richness of the task \cite{udagawa2019natural}---often use more complex strategies. Human strategies can unfold across multiple turns, \eg introducing information in installments or referring to the same dot in multiple turns without being repetitive, as A does when providing more information about the ``light grey dot`` in \Figure{fig:human-examples}a. Sophisticated strategies are also used even in single turns, \eg in \Figure{fig:human-examples}b, B's utterance ``is one on top of the other? if so pick the top one'' combines multiple types of speech act \cite{austin1962things,searle1976classification}: implicitly acknowledging A's utterance, asserting new information about the dots in view, and issuing a command.

\end{document}